\documentclass{article}

\usepackage{microtype}
\usepackage{graphicx}
\usepackage{subcaption}
\usepackage{booktabs} 

\usepackage{hyperref}

\usepackage[preprint]{icml2026}

\usepackage{amsmath}
\usepackage{amssymb}
\usepackage{mathtools}
\usepackage{amsthm}

\usepackage{multirow}
\usepackage{caption}
\usepackage{subcaption}
\usepackage{makecell}
\usepackage{graphicx}
\usepackage{colortbl}
\usepackage{tcolorbox}
\usepackage{wrapfig}
\usepackage{lipsum}
\usepackage{tablefootnote}
\usepackage{threeparttable}
\usepackage{xspace}
\usepackage{bbding}
\usepackage{xcolor}
\usepackage{fontawesome}
\usepackage{capt-of}

\usepackage[capitalize,noabbrev]{cleveref}

\theoremstyle{plain}

\theoremstyle{definition}

\theoremstyle{remark}

\usepackage[textsize=tiny]{todonotes}

\definecolor{colorV}{HTML}{03045E} 
\definecolor{colorI}{HTML}{0077B6} 
\definecolor{colorB}{HTML}{00B4D8} 
\definecolor{colorE}{HTML}{0096C7} 

\newcommand{\vibeLogo}{%
\textbf{\textcolor{colorV}{V}\textcolor{colorI}{I}\textcolor{colorB}{B}\textcolor{colorE}{E}}%
}

\usepackage{pifont}

\newcommand{\cmark}{\ding{51}}
\newcommand{\xmark}{\ding{55}}

\newcommand{\abbr}[0]{VIBE\xspace}

\newcommand{\rrparagraph}[1]{\vspace{0.5mm}\noindent\textbf{#1:}}
\newcommand{\sparagraph}[1]{\vspace{0.0mm}\noindent\textbf{#1.}}

\definecolor{best}{HTML}{f94144}
\definecolor{second}{HTML}{00B4D8}

\icmltitlerunning{
  VIBE: A Systematic Benchmark for Visual Instruction-Driven Image Editing}

\begin{document}

\twocolumn[
  \icmltitle{\textit{How Well Do Models Follow Visual Instructions?} \\
  \texorpdfstring{\vibeLogo}{VIBE}: A Systematic Benchmark for Visual Instruction-Driven Image Editing}



  \icmlsetsymbol{equal}{*}

  \begin{icmlauthorlist}
    \icmlauthor{Huanyu Zhang}{equal,ucas,casia}
    \icmlauthor{Xuehai Bai}{equal}
    \icmlauthor{Chengzu Li}{equal,cam}
    \icmlauthor{Chen Liang}{casia}
    \icmlauthor{Haochen Tian}{ucas,casia}
    \icmlauthor{Haodong Li}{scut}
    \icmlauthor{Ruichuan An}{pku}
    \icmlauthor{Yifan Zhang}{ucas,casia}
    \icmlauthor{Anna Korhonen}{cam}
    \icmlauthor{Zhang Zhang}{ucas,casia} 
    \icmlauthor{Liang Wang}{ucas,casia}
    \icmlauthor{Tieniu Tan}{nju}\\
    \icmlauthor{\textcolor{colorE}{\textit{https://vibe-benchmark.github.io/}}}{}
  \end{icmlauthorlist}

  \icmlaffiliation{ucas}{School of Artificial Intelligence, University of Chinese Academy of Science}
  \icmlaffiliation{casia}{Institute of Automation, Chinese Academy of Sciences}
  \icmlaffiliation{cam}{Language Technology Lab, University of Cambridge}
  \icmlaffiliation{pku}{Peking University}
  \icmlaffiliation{scut}{South China University of Technology}
  \icmlaffiliation{nju}{Nanjing University}

  \icmlcorrespondingauthor{Huanyu Zhang}{huanyu.zhang@cripac.ia.ac.cn}


  \vskip 0.3in
]



\printAffiliationsAndNotice{\icmlEqualContribution}

\begin{abstract}
  Recent generative models have achieved remarkable progress in image editing. 
  However, existing systems and benchmarks remain largely text-guided.
  In contrast, human communication is inherently multimodal, where visual instructions such as sketches efficiently convey spatial and structural intent.
  To address this gap, we introduce VIBE, the Visual Instruction Benchmark for Image Editing with a three-level interaction hierarchy that captures deictic grounding, morphological manipulation, and causal reasoning.
  Across these levels, we curate high-quality and diverse test cases that reflect progressively increasing complexity in visual instruction following.
  We further propose a robust LMM-as-a-judge evaluation framework with task-specific metrics to enable scalable and fine-grained assessment.
  Through a comprehensive evaluation of 17 representative open-source and proprietary image editing models, we find that proprietary models exhibit early-stage visual instruction-following capabilities and consistently outperform open-source models. 
  However, performance degrades markedly with increasing task difficulty even for the strongest systems, highlighting promising directions for future research.

\end{abstract}
\vspace{-4mm}
\section{Introduction}

Recent advances in generative models have demonstrated impressive capabilities in image editing~\cite{bananapro,seedream2025seedream}.
However, most existing systems remain predominantly text-guided~\cite{huang2025diffusion,liu2025step1x}, a paradigm that imposes a significant double-sided cognitive burden.
For the user, conveying precise spatial or structural intent through text alone is often a cumbersome task, necessitating verbose and pedantic descriptions. 
Meanwhile, these dense textual instructions are equally demanding for the models to understand without ambiguity and reconstruct into accurate spatial intent \citep{li-etal-2024-topviewrs}. 

\begin{figure}
    \centering
    \includegraphics[width=0.95\linewidth]{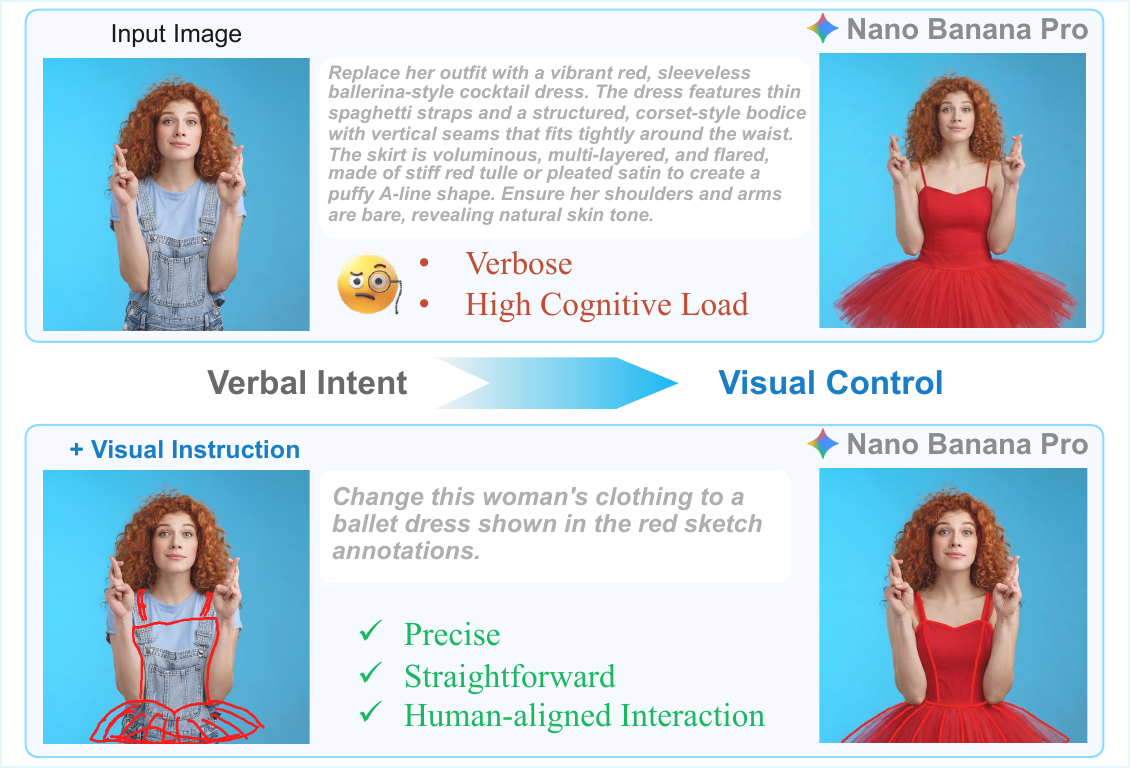}
    \vspace{-2mm}
\caption{\textbf{Motivation and scope of the \abbr benchmark.} 
Traditional image editing is largely text-guided, where conveying spatial intent relies on verbose descriptions and incurs high cognitive load.
In contrast, visual instructions enable precise and explicit grounding, providing a more human-aligned interaction paradigm.
VIBE is designed to fill the evaluation gap by systematically benchmarking this visual intruction-guided multi-modal image editing.
}
    \label{fig:intro}
    \vspace{-6mm}
\end{figure}

In contrast, human communication is inherently multimodal: users naturally combine language with visual cues, such as sketches, arrows, or region annotations, to disambiguate intent and achieve precise control. 
These \textbf{visual instructions} offer a more natural and efficient interaction paradigm, enabling explicitly grounded editing that aligns with how humans intuitively reason about visual content~\cite{buxton2010sketching,tversky2013visualizing}.
While state-of-the-art models like Nano Banana Pro are beginning to follow these intuitive cues (see \figurename~\ref{fig:intro}), existing benchmarks~\cite{zhang2023magicbrush,zhao2025risebench,wu2025kris} are still limited to text-only guidance, failing to capture the efficiency and clarity of multimodal interaction.

To address this gap, we introduce \textbf{\abbr}, the \textbf{V}isual \textbf{I}nstruction \textbf{B}enchmark for Image \textbf{E}diting, which is designed to systematically evaluate image editing guided by visual instructions.
\abbr formalizes visual instructions as spatially-anchored cues, such as sketches or manipulative vectors, that provide the geometric grounding necessary to resolve the ambiguities inherent in text-only instructions.
To evaluate these capabilities, we structure \abbr along a three-level interaction hierarchy that represents a progression in communicative and reasoning complexity:
(1) \textit{Deictic} grounding to identify and isolate sptaial lcoations; 
(2) \textit{Morphological} manipulation to specify shape, pose and geometric properties;
(3) \textit{Causal} reasoning to predict the logical visual consequence of a specified action through manipulative vectors. 
Across this hierarchy, \abbr comprises 10 functionally diverse subtasks, capturing a wide range of behaviors from simple attribute swapping to complex structural synthesis.
In total, \abbr contains 1,034 samples, all of which are manually annotated or carefully verified by human.
For evaluation, we develop specialized evaluation metrics tailored to the objectives of each task, and adopt an LMM-as-a-judge framework to assess instruction-following behavior.
To verify the reliability and effectiveness of the proposed evaluation framework, we further perform extensive experiments, observing high correlation between LMM-based judgment and human expert assessment. 

Through extensive benchmarking across 10 open-source and 7 proprietary models, our experiments yield three key findings.
First, frontier proprietary models demonstrate emerging visual instruction-following capabilities, 
indicating reliable spatial grounding under explicit visual instructions.
Second, substantial performance gaps persist between proprietary and open-source models across all levels, with proprietary systems consistently achieving higher scores.
Third, across all proprietary models, performance degrades from the \textit{Deictic} level to the \textit{Causal} level.
Even the strongest models achieve average scores below 50\% on the \textit{Causal} level, indicating that complex causal reasoning remains a significant challenge.
This consistent performance degradation further validates the hierarchical design of \abbr.

In summary, our main contributions are as follows:
\vspace{-4mm}  
\begin{itemize}
\setlength{\itemsep}{0pt} 
    \setlength{\parskip}{0pt}
    \setlength{\topsep}{0pt}
\item We introduce \abbr, the first benchmark for systematically evaluating visual instruction-guided image editing, establishing a comprehensive framework for assessing multimodal instruction-following.
\item We formulate a cognitively motivated hierarchy spanning deictic grounding, morphological manipulation, and causal reasoning, and design task-specific evaluation metrics supported by a validated LMM-as-a-judge framework for scalable and reliable assessment.
\item We conduct a comprehensive evaluation of 17 models, revealing clear capability gaps and offering new insights into the strengths, limitations, and future challenges of visual instruction-guided image editing.
\end{itemize}

\vspace{-3mm}
\section{VIBE}
\label{sec:vibe_benchmark}
\vspace{-1mm}

To bridge the gap between linguistic instructions and precise image manipulation, we introduce the \textbf{VIBE} benchmark.
In this section, we first formalize the concept of visual instructions. 
Then, we present the hierarchical task suites within VIBE, ranging from deictic grounding to causal reasoning tasks involving physical interactions. 
Finally, we detail our data construction pipeline and task-specific metrics developed to quantify the precision of multi-modal instruction-following.

\subsection{Visual Instruction Formulation}
\label{sec:formulation}
In human communication, visual cues are often used to disambiguate intent and anchor meaning in space \citep{herring2015new}.
Motivated by this, we formalize \textit{visual instructions} as spatially explicit signals that provide direct grounding constraints for image editing.
Unlike textual instructions, which require the model to translate abstract linguistic symbols into spatial coordinates, a visual instruction acts as a direct geometric interface, bridging the gap between high-level intent and pixel-level execution.
Given a source image $I_{in}$, a textual instruction $T$, and a visual instruction $V$, provided either as a separate image or as overlaid annotations, a model $\phi$ is required to generate an edited output $I_{out}$ as:
\begin{equation}
    I_{out} = \phi (I_{in}, T, V).
\end{equation}

\subsection{Benchmark Construction}
\label{sec:construction}

\begin{figure}
    \centering
    \includegraphics[width=\linewidth]{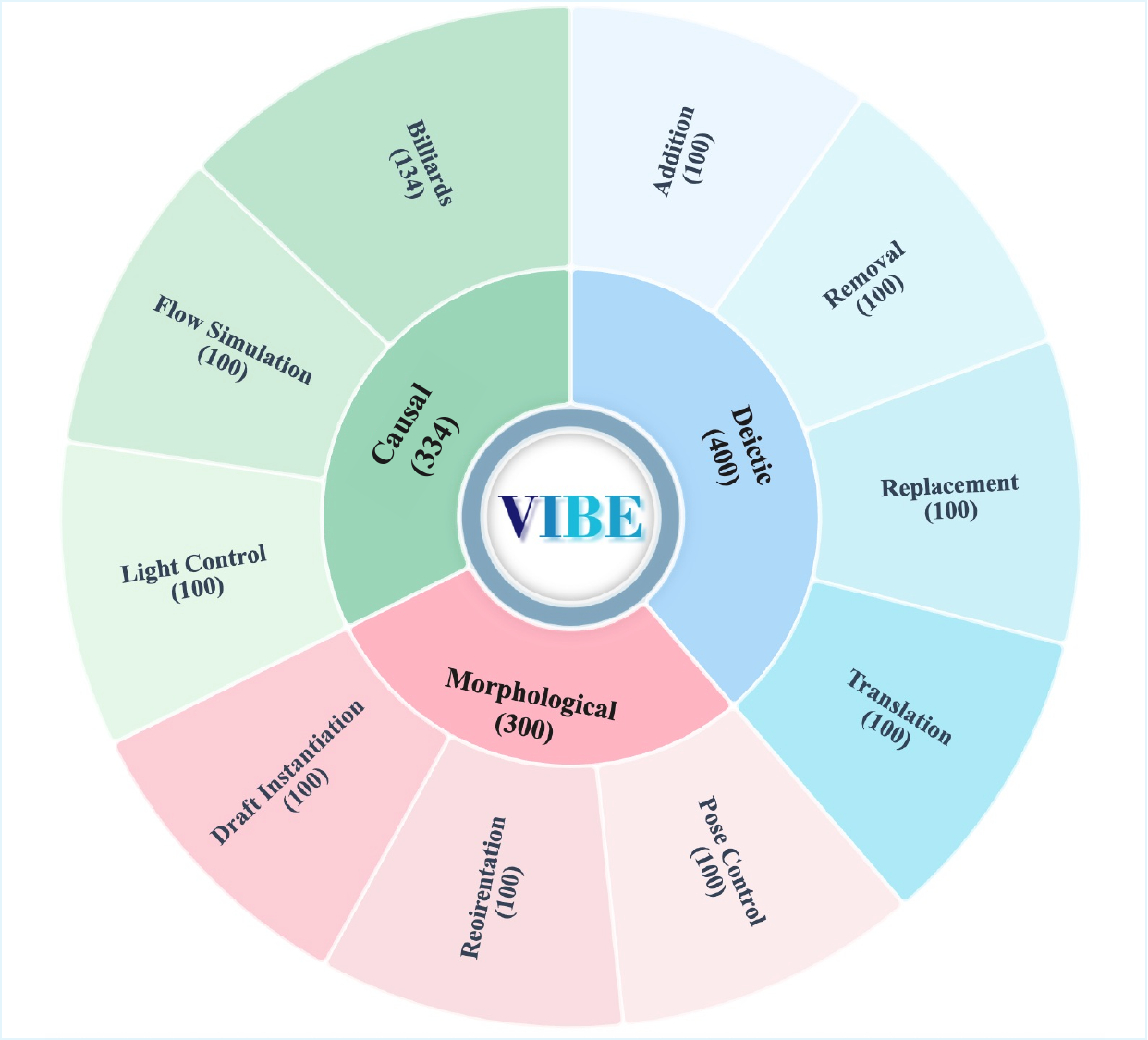}
\caption{\textbf{Composition of \abbr.} \abbr comprises 1,034 samples across 10 tasks, organized into a three-level hierarchy that reflects increasing interaction and reasoning complexity, from deictic grounding and morphological manipulation to causal reasoning.}
    \label{fig:tasks}
    \vspace{-3mm}
\end{figure}

\begin{figure*}[t]
    \centering
    \includegraphics[width=\linewidth]{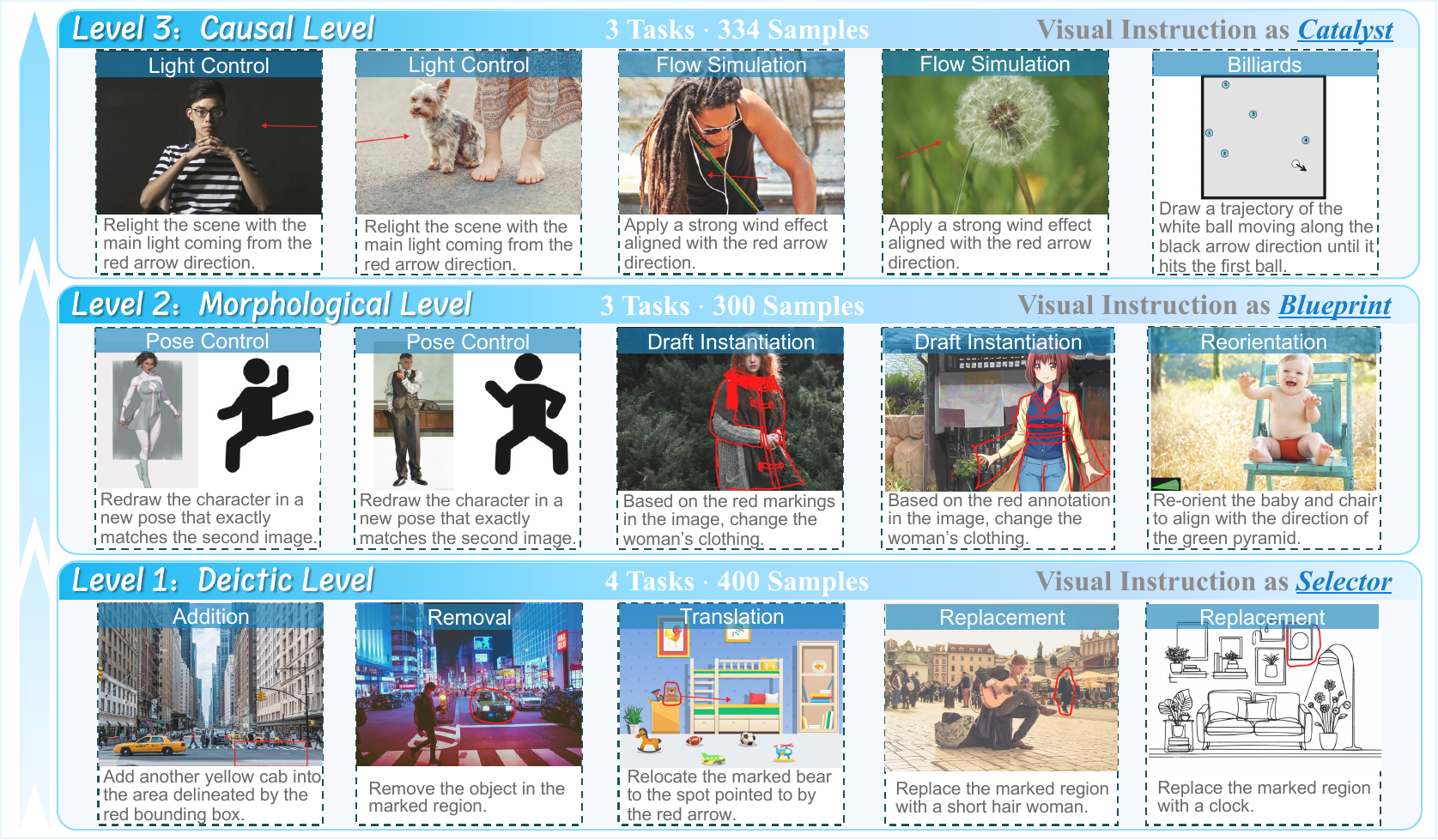}
    \vspace{-2mm}
\caption{\textbf{Overview of \abbr.}
\abbr organizes visual instruction-guided image editing into a three-level interaction hierarchy with increasing task complexity.
The \textit{Deictic} Level treats visual instructions as selectors that specify localized regions or objects for basic spatial operations. 
The \textit{Morphological} Level interprets visual instructions as blueprints that define abstract structural constraints. 
The \textit{Causal} Level views visual instructions as catalysts that encode underlying physical or logical dynamics.
}
    \label{fig:overview}
    \vspace{-3mm}
\end{figure*}

Based on how models interpret and execute visual instructions, we organize all tasks in VIBE into a three-level interaction hierarchy.
As demonstrated in \figurename~\ref{fig:overview}, each level reflects an increasing degree of abstraction and reasoning complexity, ranging from spatial grounding to structural realization and causal simulation.

\rrparagraph{Level 1: \textit{Deictic} Level} 
At this level, the instruction acts as a \textit{selector}. 
Markers like bounding boxes or arrows serve as digital deictic cues, indicating the spatial location of an intended edit. 
This level primarily evaluates a model’s spatial grounding and basic visual awareness.
As illustrated in \figurename~\ref{fig:tasks}, there are four fundamental editing tasks including \textit{\textbf{Addition}}, \textit{\textbf{Removal}}, \textit{\textbf{Replacement}}, and \textit{\textbf{Translation}}.

Specifically, Addition (AD) requires the model to introduce new visual content confined to the designated region, while preserving all pre-existing elements elsewhere.
Removal (RM) instructs the model to eliminate the target object or region and plausibly reconstruct the underlying background without leaving residual artifacts.
Replacement (RP) involves substituting the content within the specified region with a different object, while maintaining the original spatial extent and placement.
Translation (TR) focuses on repositioning a selected object according to spatial cues, without modifying its visual appearance, structure, or identity.

All four tasks share the same set of source images to ensure a fair comparison across different editing operations.
Besides, to ensure broad coverage of real-world and creative scenarios, we include images spanning diverse visual styles.
Specifically, each task contains 100 samples, with a balanced distribution of visual styles, including 34 real-world images, 33 animated images, and 33 sketch-based illustrations.
And all visual instructions and annotations for the \textit{Deictic} Level are manually created, guaranteeing precise spatial grounding and unambiguous task intent.

\rrparagraph{Level 2: \textit{Morphological} Level} 
At the \textit{Morphological} Level, visual instructions function as \textit{blueprints}.
Sparse representations, including skeletons or schematic sketches, specify the structural constraints of the target transformation.  
The model is required to map these abstract forms into coherent, style-consistent geometries and appearances.
As shown in \figurename~\ref{fig:tasks}, we instantiate this level with three tasks: \textit{\textbf{Pose Control}}, \textit{\textbf{Reorientation}}, and \textit{\textbf{Draft Instantiation}}.

Pose Control (PC) requires the model to transform a character in the input image to exactly match the pose provided by a reference image, while preserving the character's identity, appearance, and visual style.
Reorientation (RO) focuses on aligning the orientation of an object with a given reference viewing frustum or directional cue.
Draft Instantiation (DI) challenges the model to convert sketch-based annotations overlaid on the input image into a fully realized output, while maintaining consistency with the original scene in both content and style. 

For the \textit{Morphological} Level, each task is also constructed using 100 distinct samples.
In particular, for Draft Instantiation, the visual instructions consist of hand-drawn sketches directly overlaid on the input images, ensuring faithful representation of abstract structural intent.
For Pose Control, each sample is constructed from a manually curated pair of images, consisting of an input image and a reference image that specifies the target pose.
For Reorientation, the reference viewing frustums or directional cues are individually annotated by human annotators.
These annotations precisely define the intended object orientation and ensure unambiguous structural constraints for evaluation.

\rrparagraph{Level 3: \textit{Causal} Level} 
The \textit{Causal} Level represents the highest level of interaction in our hierarchy.
At this level, visual instructions operate as \textit{catalysts}.
Visual cues such as force vectors or motion arrows do not depict the final outcome but instead encode the underlying causal dynamics to be applied.
This requires models to possess an internal world model capable of predicting the logical outcome of physical events.
As illustrated in \figurename~\ref{fig:tasks}, this level contains three challenging tasks: \textit{\textbf{Light Control}}, \textit{\textbf{Flow Simulation}}, and \textit{\textbf{Billiards}}.

Light Control (LC) requires the model to modify the direction of illumination according to annotated arrows, updating shading, shadows, and highlights consistently with the new lighting direction.
Flow Simulation (FS) instructs the model to simulate wind flow based on directional cues.
Objects in the scene should respond plausibly to the implied airflow, exhibiting appropriate deformation or motion.
Billiards (BI) challenges the model to predict the trajectory of a ball under an applied force indicated by arrows.
The task requires reasoning about interactions with the environment, including collisions and subsequent rebounds.

In the \textit{Causal} Level, both \textit{Flow Simulation} and \textit{Light Control} comprise 100 samples each, where all causal cues and annotations are manually specified to ensure physical plausibility and consistency with common-sense dynamics.
The \textit{Billiards} task is constructed using a hybrid approach.
We first synthesize 200 candidate cases by scripts, covering scenarios with increasing difficulty ranging from two to seven collisions.
These candidates are then manually inspected, resulting in a final set of 134 high-quality samples.

To maintain high annotation quality, all samples undergo multiple rounds of manual verification.
More details about data collection and annotation are provided in Appendix~\ref{app:data}.

\subsection{Evaluation Pipeline}
\label{sec:eval}

Evaluating visual instruction-guided image editing remains challenging due to the open-ended nature of visual outputs and the absence of a unique ground truth.
Recent studies~\cite{zhao2025envisioning,GPT5,gemini} have shown that large multimodal models (LMMs) exhibit strong visual reasoning and alignment capabilities, making them suitable as automated evaluators.
Following this line of work, we adopt an LMM-as-a-Judge evaluation paradigm.
Specifically, we employ a frontier LMM (GPT5.1\footnote{We use the 2025-11-13 version of GPT5.1 via the Azure API.}) as the evaluator.
For each sample, the evaluator is provided with the input image, the textual instruction, the corresponding visual instruction, and the generated output.
The evaluator is required to determine whether the generated result correctly fulfills the specified instruction.

Evaluation criteria are designed in a task-specific manner to reflect the distinct requirements of different tasks.
For tasks in the \textit{Deictic} Level, the evaluation is conducted along three complementary criteria that jointly capture instruction compliance, locality preservation, and visual quality:

\begin{itemize}
    \setlength{\itemsep}{0pt} 
    \setlength{\parskip}{0pt}
    \setlength{\topsep}{0pt}  
    \item \textbf{Instruction Adherence ($\mathcal{IA}$)} measures whether the model faithfully given the specified visual instruction.
    It is defined as the conjunction of three binary criteria: 
    (i) \emph{visual instruction localization correctness}, which verifies whether the model correctly identifies the target region specified by the visual cues.; 
    (ii) \emph{visual operator type compliance}, which checks whether the applied editing operation matches the specified visual operator; 
    and (iii) \emph{textual action semantic compliance}, which evaluates consistency with the textual instruction.
    $\mathcal{IA}$ is computed as the average of the three scores.

    \item \textbf{Contextual Preservation ($\mathcal{CP}$)} evaluates whether non-target regions remain intact after editing.
    It is assigned a binary score based on whether unintended modifications to background elements, surrounding objects, or the global scene structure are presented, or not.

    \item \textbf{Visual Coherence ($\mathcal{VC}$)} measures the perceptual integrity of the edited result.
    It is defined through three binary sub-criteria: (i) \emph{style consistency}, assessing whether the edited image conforms to the artistic style of the source image; (ii) \emph{visual seamlessness}, assessing whether the edited region integrates smoothly with its surroundings; and (iii) \emph{artifact-free generation}, assessing the absence of visual artifacts such as blurring, distortion, or unnatural seams.
    $\mathcal{VC}$ is computed as the average of the three scores.

\end{itemize}

\begin{table*}[t]
\caption{ \textbf{Experimental results on \abbr.} We report task-wise and overall performance across the \textit{Deictic}, \textit{Morphological}, and \textit{Causal} levels. The best and second-best results are highlighted in \textcolor{best}{red} and \textcolor{second}{blue}, respectively.
}
\vspace{-2mm}
\label{tab:main}

\rowcolors{4}{white}{gray!15}
\renewcommand{\arraystretch}{1.2}
\newcolumntype{C}{>{\centering\arraybackslash}p{0.08\textwidth}}
\newcolumntype{E}{>{\centering\arraybackslash}p{0.07\textwidth}}
\resizebox{\linewidth}{!}{
\begin{tabular}{l|C|cccc|c|ccc|c|ccc|c|c}
\hline
\multirow{2}{*}{Model}       & \multirow{2}{*}{Multi-Img} & \multicolumn{5}{c|}{\textbf{\textit{Deictic} Level}}                                                                                                          & \multicolumn{4}{c|}{\textbf{\textit{Morphological} Level}}                                                                 & \multicolumn{4}{c|}{\textbf{\textit{Causal} Level}}                                                         & \multirow{2}{*}{\textbf{Overall}} \\ \cline{3-15}
                             &                            & {AD} & {RM} & {RP} & {TR} & {\textbf{Avg}} & PC             & {RO} & {DI}                  & {\textbf{Avg}} & LC            & FS                & {BI} & {\textbf{Avg}} &                          \\
                            \hline
                             
Nano Banana Pro              & \cmark                        & {\textcolor{best}{82.17}}            & {94.07}           & \textcolor{best}{88.26}               & \textcolor{best}{74.80}               & \textcolor{best}{84.83}       & \textcolor{best}{72.33}   & \textcolor{best}{36.04}                 & \textcolor{best}{88.02}         & \textcolor{best}{65.46}       & \textcolor{best}{60.34}   & \textcolor{best}{59.25}      & \textcolor{best}{15.92}             & \textcolor{best}{45.17}       & \textcolor{best}{65.15}           \\
Nano Banana                  & \cmark                        & \textcolor{second}{81.34}                     & 93.50                    & 79.05                        & 46.53                        & 75.11                & \textcolor{second}{67.71}            & 33.45                          & \textcolor{second}{85.60}                  & \textcolor{second}{62.25}                & 34.75            & \textcolor{second}{52.64}               & 1.87                       & 29.75                & \textcolor{second}{55.70}                    \\
GPT-image-1                  & \cmark                        & 55.61                     & 69.00                    & 62.63                        & 47.00                        & 58.56                & 64.39            & 11.09                          & 77.32                  & 50.93                & 25.18            & 39.48               & 4.73                       & 23.13                & 44.21                    \\
Seedream 4.5                 & \cmark                        & 81.24                     & \textcolor{second}{95.82}                    & \textcolor{second}{81.82}                        & \textcolor{second}{48.93}                        & \textcolor{second}{76.95}                & 66.79            & 20.11                          & 82.33                  & 56.41                & \textcolor{second}{50.50}            & 45.55               & 2.99                       & 33.01                & 55.46                    \\
Seedream 4.0                 & \cmark                        & 74.02                     & 93.04                    & 79.35                        & 33.29                        & 69.93                & 58.27            & 30.37                          & 72.09                  & 53.58                & 47.00            & 43.59               & 4.11                       & 31.57                & 51.69                    \\
Wan 2.6                       & \cmark                        & 66.01                     & 92.90                    & 68.15                        & 40.95                        & 67.00                & 59.66            & \textcolor{second}{34.89}                          & 80.23                  & 58.26                & 44.46            & 50.79               & \textcolor{second}{9.08}                       & \textcolor{second}{34.78}                & 53.35                    \\ 
Wan 2.5                       & \cmark                        & 73.59                     & \textcolor{best}{96.90}                    & 76.99                        & 36.80                        & 71.07                & 55.76            & 25.77                          & 78.78                  & 53.44                & 33.33            & 51.98               & 7.84                       & 31.05                & 51.85                    \\ \hline
FLUX2-dev                    & \cmark                        & 64.57                     & 8.00                     & 54.40                        & 5.58                         & 33.14                & 28.76            & 22.77                          & 60.68                  & 37.40                & 33.74            & 30.81               & 2.36                       & 22.30                & 30.95                    \\
Qwen-Image-Edit-2509         & \cmark                        & 55.28                     & 14.38                    & 30.13                        & 14.48                        & 28.57                & 15.14            & 17.67                          & 21.38                  & 18.06                & 28.00            & 44.40               & 2.36                       & 24.92                & 23.85                    \\
Qwen-Image-Edit              & \xmark                         & 44.20                     & 24.88                    & 30.48                        & 11.11                        & 27.67                & -                & 21.33                          & 54.65                  & -                    & 22.00            & 32.38               & 3.11                       & 19.16                & 23.42                    \\
Edit-R1-Qwen-Image-Edit & \cmark                        & 56.77                     & 4.86                     & 29.47                        & 11.33                        & 25.61                & 16.23            & 20.27                          & 15.42                  & 17.31                & 25.67            & 40.22               & 2.24                       & 22.71                & 21.87                    \\
BAGEL-think                  & \cmark                        & 40.44                     & 14.59                    & 35.38                        & 14.46                        & 26.22                & 8.50             & 23.60                          & 50.34                  & 27.48                & 21.17            & 28.04               & 5.22                       & 18.14                & 23.95                    \\
BAGEL                        & \cmark                        & 33.87                     & 11.33                    & 29.26                        & 14.05                        & 18.21                & 7.61             & 28.05                          & 48.23                  & 27.96                & 21.57            & 33.63               & 5.97                       & 20.39                & 22.19                    \\
Step1X-Edit-v1p2             & \xmark                         & 33.92                     & 12.59                    & 28.17                        & 13.52                        & 22.05                & -                & 25.17                          & 71.48                  & -                    & 25.00            & 29.67               & 0.37                       & 18.35                & 20.20                    \\
OmniGen2                     & \cmark                        & 26.29                     & 26.20                    & 20.84                        & 4.51                         & 19.46                & 11.40            & 17.44                          & 30.51                  & 19.78                & 17.33            & 17.61               & 2.74                       & 12.56                & 17.27                    \\
UniWorld-V1                  & \cmark                        & 15.18                     & 14.52                    & 22.03                        & 3.59                         & 13.83                & 11.95            & 16.57                          & 34.43                  & 20.98                & 15.50            & 9.28                & 0.00                       & 8.26                 & 14.36                    \\
OmniGen                      & \cmark                        & 2.63                      & 7.48                     & 5.26                         & 1.23                         & 4.15                 & 5.79             & 13.88                          & 3.93                   & 7.87                 & 2.33             & 2.00                & 0.00                       & 1.44                 & 4.49                     \\ \hline
\end{tabular}
}
\vspace{-4mm}
\end{table*}

Instruction Adherence serves as a prerequisite for meaningful visual evaluation.
If $\mathcal{IA}$ equals zero, $\mathcal{VC}$ is automatically set to zero.
Similar to VIEScore~\cite{ku2024viescore}, the final score for each sample is computed as the geometric mean of the three criteria:
\begin{equation}
\text{Score} = (\mathcal{IA} \cdot \mathcal{CP} \cdot \mathcal{VC})^{\frac{1}{3}}.
\end{equation}
This formulation ensures that high scores are assigned only when correct instruction execution, strict context preservation, and coherent visual synthesis are simultaneously achieved.
To improve scoring reliability and reduce evaluator ambiguity, we design the majority of sub-metrics as binary decisions, encouraging the evaluator to focus on clear-cut judgments.
Evaluation metrics for other tasks are designed following the similar principles as those of the \textit{Deictic} Level, and detailed definitions as well as the exact evaluation prompts are provided in the Appendix~\ref{app:eval}.

\section{Experiments}

To ensure a comprehensive evaluation, we benchmarked a total of 17 models, categorized into proprietary and open-source systems. 
The proprietary category includes leading commercial models: Nano Banana~\cite{banana} and its Pro variant~\cite{bananapro}, GPT-image-1~\cite{hurst2024gpt}, the Seedream series (4.0 and 4.5)~\cite{seedream2025seedream}, and the Wan series (2.5 and 2.6)~\cite{wan}. 
Besides, the open-source models comprise FLUX2-dev~\cite{blackforestlabs2025flux2}, Qwen-Image-Edit-2509, Qwen-Image-Edit~\cite{wu2025qwen}, 
Edit-R1-Qwen-Image-Edit-2509~\cite{li2025editr1}, Bagel~\cite{deng2025bagel}, Step1X-Edit-v1p2 (think + reflection version)~\cite{liu2025step1x}, UniWorld-V1~\cite{lin2025uniworld}, and the OmniGen series (OmniGen~\cite{xiao2025omnigen} and OmniGen2~\cite{wu2025omnigen2}).

As discussed in Section~\ref{sec:eval}, we employ GPT-5.1 as the evaluator in our evaluation framework.
To reduce stochastic variance and improve score stability, each sample is evaluated three times independently, and the final score is reported as the average over the three runs.

\subsection{Main Results on VIBE}

We report the performance score on a 100-point scale in Table~\ref{tab:main}, covering task-wise results, level-wise averages, and overall performance across all evaluated models.
These results provide a comprehensive assessment of model capabilities under progressively more demanding visual instruction settings.
The results reveal several consistent performance trends across interaction levels and model categories.

\sparagraph{Proprietary models exhibit early-stage visual instruction-following capabilities}
Specifically, nearly all proprietary models achieve scores above 60 on Addition, Removal, and Replacement tasks in the \textit{Deictic} Level, suggesting reliable performance on explicit, region-based visual instructions.
These models also perform well on more fine-grained visual instruction tasks, such as Pose Control and Draft Instantiation, which require structured manipulation guided by reference images or sketches.
In contrast, performance is comparatively moderate on tasks involving more abstract directional or referential visual instructions, including Translation and Reorientation.
These tasks demand reasoning over spatial relations and orientation changes, which are less directly grounded by localized visual cues.
Together, these results suggest that while proprietary models have begun to acquire foundational visual instruction-following capabilities, challenges remain in handling abstract, direction-based instructions that require higher-level spatial reasoning.

\begin{figure*}
    \centering
    \includegraphics[width=\linewidth]{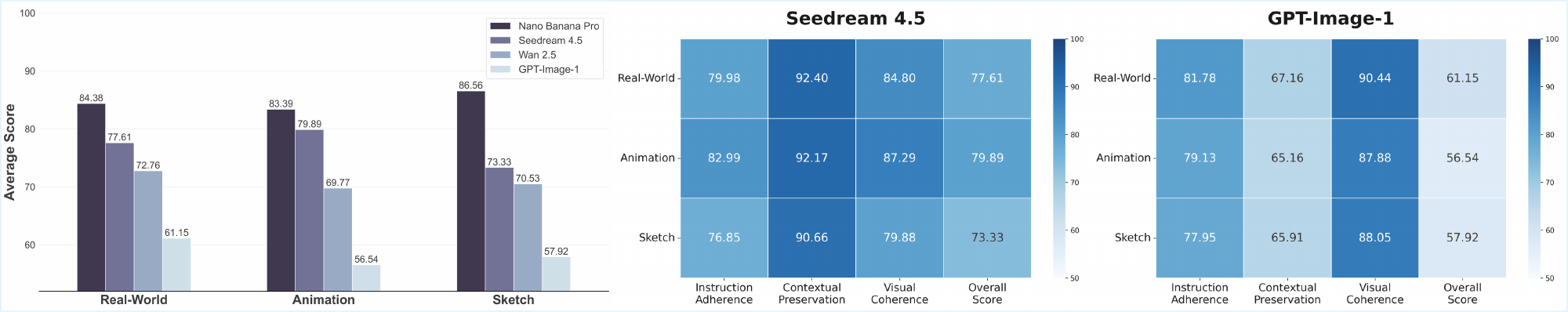}
    \vspace{-5mm}
\caption{\textbf{Performance across image styles on the Deictic Level.} Left: Average Deictic Level scores across real-world, animation, and sketch images for four proprietary models.
Right: Metric-level heatmaps for Seedream~4.5 and GPT-Image-1, illustrating style-dependent variations in Instruction Adherence, Contextual Preservation, and Visual Coherence.}
    \label{fig:style}
    \vspace{-3mm}
\end{figure*}

\sparagraph{Substantial performance gaps persist between proprietary and open-source models across all levels}
As shown in Table~\ref{tab:main}, proprietary systems achieve substantially higher overall scores than open-source models.
This performance gap reflects the advantages of proprietary models in terms of model scale and training data diversity, which contribute to more accurate interpretation and execution of visual instructions across all levels.
At the same time, these results highlight a clear opportunity for future open-source research to narrow this gap by improving instruction understanding, multimodal alignment, and higher-level reasoning capabilities.

\sparagraph{Performance on proprietary models degrades from the \textit{Deictic} Level to the \textit{Causal} Level}
When results are aggregated at different task levels, proprietary models exhibit a clear performance degradation from the Deictic Level to the Morphological Level, and further to the Causal Level.
The observed degradation reflects the increasing interaction complexity, where higher levels require not only localized editing but also structural abstraction and causal reasoning.
These results therefore validate the hierarchical design of \abbr, demonstrating a systematic performance decline as models are challenged with progressively more demanding forms of visual instruction.
Detailed error analyses across tasks and models are provided in Appendix~\ref{app:exp}.

\section{Discussion and Analysis}

\begin{figure}[t]
    \centering
    \includegraphics[width=0.95\linewidth]{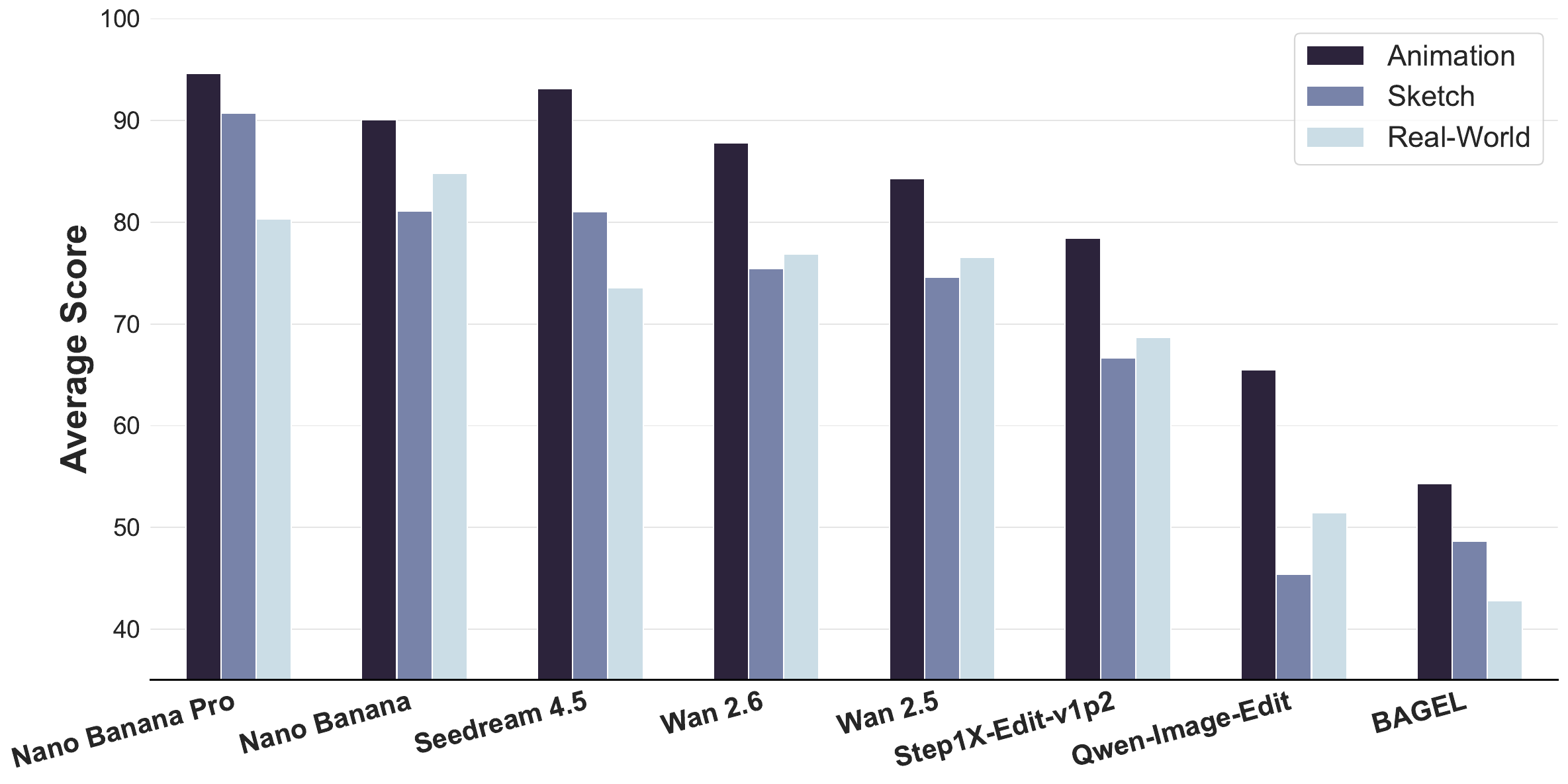}
    \vspace{-2mm}
\caption{\textbf{Style-wise performance on Draft Instantiation.}}
    \label{fig:draft_style}
    \vspace{-4mm}
\end{figure}

\subsection{Style-wise Performance Analysis}

As described in Section~\ref{sec:construction}, the four \textit{Deictic} Level tasks share a common set of 100 source images evenly distributed across real-world, animation, and sketch styles.
We select four proprietary models from different organizations, including Nano Banana Pro, Seedream 4.5, Wan 2.5 and GPT-Image-1, and report their average \textit{Deictic} Level scores across styles in the left chart of \figurename~\ref{fig:style}.

We find that \textbf{models exhibit distinct and model-specific preferences across visual styles on \textit{Deictic} Level tasks.}
As shown in Figure~\ref{fig:style}, Nano Banana Pro and Wan~2.5 display relatively balanced performance across all three styles.
In contrast, Seedream~4.5 shows a pronounced performance degradation on sketch-style images, whereas GPT-Image-1 shows the opposite tendency, achieving its strongest results on real-world images.
Such differences are likely attributable to variations in training data composition and representation biases across models.

To further investigate the source of the observed style preferences, we conduct a more fine-grained analysis at the metric level.
Specifically, we visualize the performance of Seedream~4.5 and GPT-Image-1 on individual evaluation metrics using the heatmap shown on the right side of Figure~\ref{fig:style}.
For Seedream~4.5, the degradation on sketch-style images can be primarily attributed to a substantial drop in \textit{Instruction Adherence} and \textit{Visual Coherence}, which are markedly lower than those on real-world and animation images.
Qualitative examples in Figure~\ref{fig:seed_style} further illustrate this behavior, where it often fails to generate objects that are stylistically consistent within the sketch domain.
In contrast, GPT-Image-1 shows a clear preference for real-world images, where it outperforms animation and sketch inputs across all evaluation metrics.
As a result, its overall scores are highest on real-world inputs, as shown in Figure~\ref{fig:style}.

In addition, we analyze style-wise performance on the \textit{Draft Instantiation} task, as illustrated in \figurename~\ref{fig:draft_style}.
Unlike \textit{Deictic} Level tasks, this task exhibits a clearer style preference, where most models achieve notably higher performance on animation-style images.
This trend suggests that animated images, which often feature cleaner contours and more explicit structural cues, may better align with sketch-based visual instructions.
Complete quantitative results and representative examples are provided in the Appendix~\ref{app:di_style}.

\begin{table}[]
\caption{\textbf{Single-task vs. Multi-task visual instruction following.} Quantitative results of five proprietary models evaluated on single-task, double-task, and triple-task compositions of Deictic Level.}
\label{tab:multi-tasks}
\vspace{-2mm}
\renewcommand{\arraystretch}{1.3}
\resizebox{0.95\linewidth}{!}{
\begin{tabular}{l|ccc}
\hline
                & Single Task & Double Tasks & Triple Tasks \\ \hline
Nano Banana Pro & 84.83       & 80.22       & 75.48       \\ \hline
Nano Banana     & 75.11       & 65.66       & 66.77       \\ \hline
GPT-image-1     & 58.56       & 47.32       & 42.85       \\ \hline
Seedream 4.5    & 76.95       & 60.16       & 61.50       \\ \hline
Seedream 4.0    & 69.93       & 62.62       & 49.15       \\ \hline
\end{tabular}
}
\vspace{-3mm}
\end{table}

\begin{figure*}
    \centering
    \includegraphics[width=\linewidth]{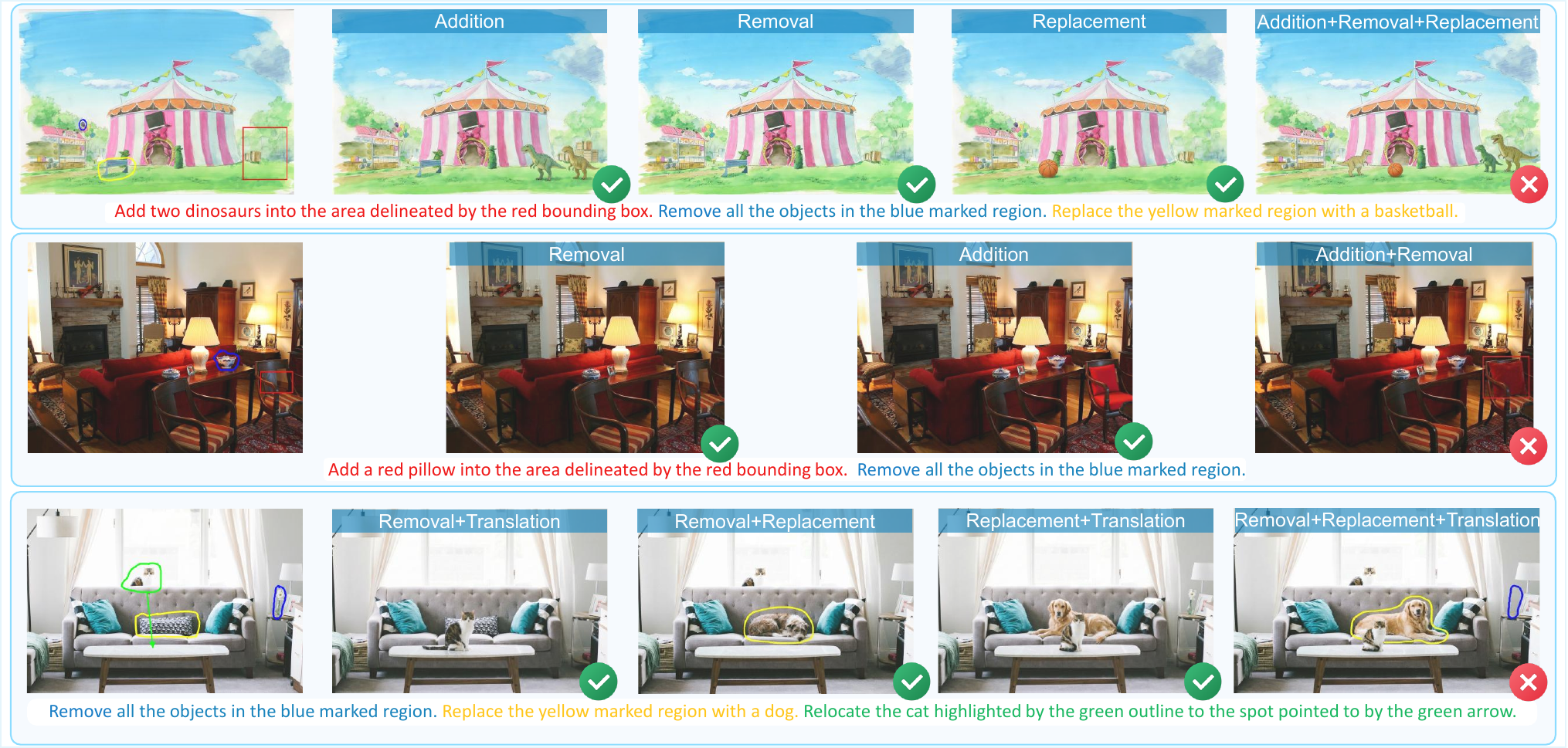}
    \vspace{-4mm}
\caption{\textbf{Qualitative examples of multi-task visual instruction following. }Examples illustrating model behavior under composed visual instructions.
In the first two rows, models correctly execute individual instructions in isolation but fail when the same instructions are combined.
The third row shows a case where a model succeeds with two instructions but fails when an additional instruction is introduced.}
    \label{fig:multitask}
    \vspace{-2mm}
\end{figure*}

\begin{figure}[t]
    \centering
    \includegraphics[width=0.84\linewidth]{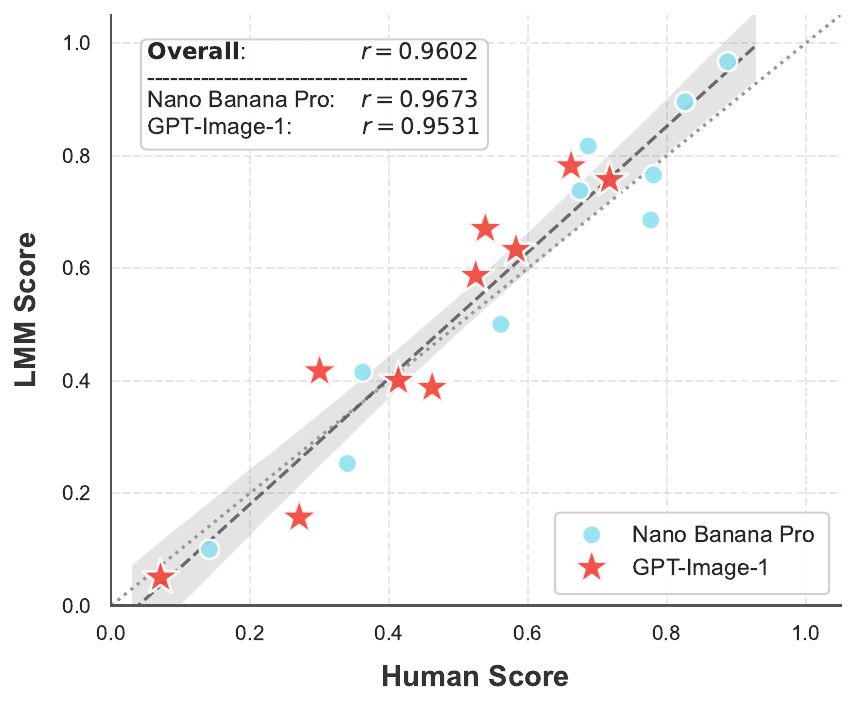}
    \vspace{-2mm}
\caption{{Pearson correlation between human expert scores and LMM-based evaluation scores for Nano Banana Pro and GPT-Image-1, demonstrating a strong alignment between human judgments and the LMM-as-a-Judge evaluator.}}
    \label{fig:human}
    \vspace{-4mm}
\end{figure}

\begin{figure*}
    \centering
    \includegraphics[width=\linewidth]{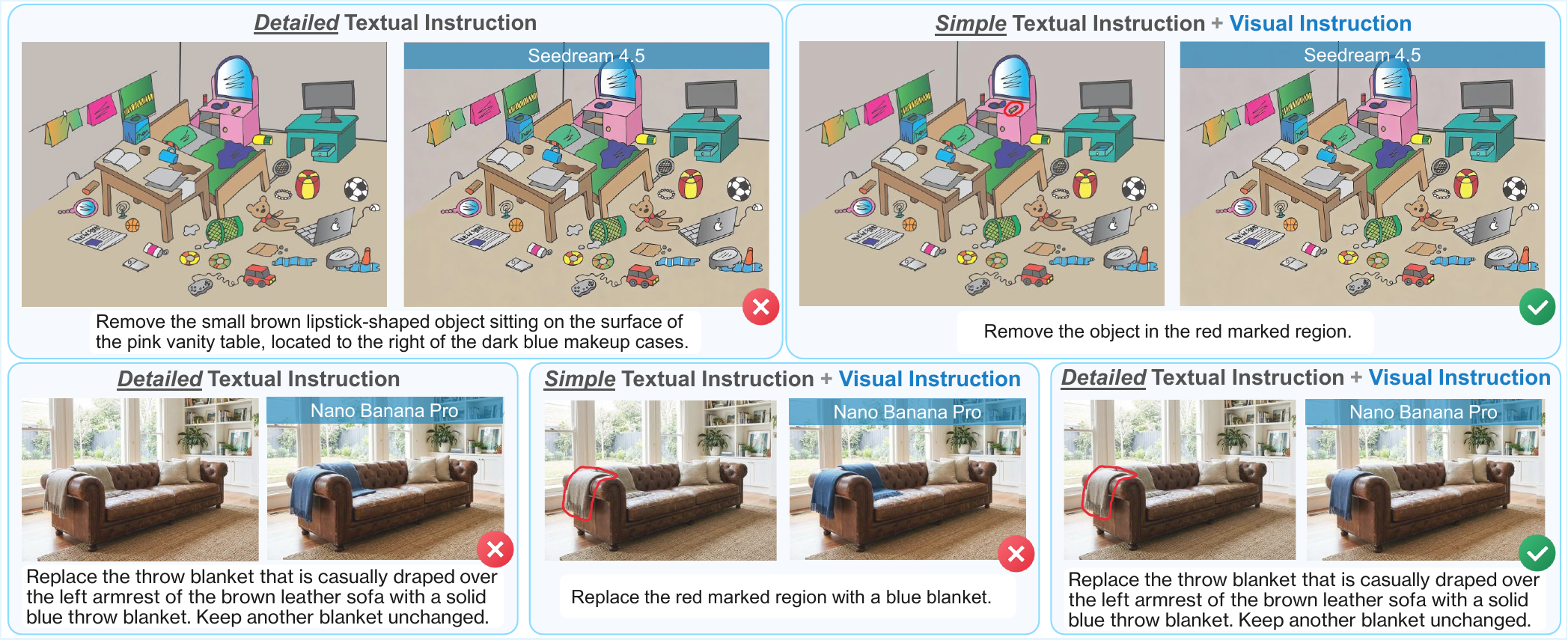}
\caption{Two representative cases illustrating how textual and visual instructions interact.
The first case shows that visual instructions can resolve target ambiguity that detailed text alone fails to address.
The second case demonstrates that complex semantic constraints require the joint use of detailed textual and visual instructions.
}
    \label{fig:instruction}
\end{figure*}

\subsection{Multi-task Visual Instruction Following}

Motivated by the observation that proprietary models exhibit early-stage visual instruction-following capabilities on individual \textit{Deictic} Level tasks, we further explore their performance under more demanding multi-task settings.
Specifically, we examine whether models can follow multiple visual instructions within a single query by constructing multi-task evaluations based on \textit{Deictic} Level tasks.
We form 6 two-task combinations by pairing all four tasks, and 4 three-task combinations by further composing them.
To ensure comparability with the single-task setting, images are curated with balanced visual styles.
For each task combination, we select five images per style (real-world, animation, and sketch), yielding 15 images per combination.
Visual instructions are manually verified to ensure that different tasks do not overlap spatially or semantically.
In total, this process yields 90 double-task cases and 60 triple-task cases.

We evaluate five proprietary models, as reported in Table~\ref{tab:multi-tasks}.
Across all models, performance exhibits a clear drop when moving from single-task to multi-task settings.
Figure~\ref{fig:multitask} provides qualitative examples illustrating this behavior, where models successfully execute individual visual instructions in isolation but fail to correctly compose multiple instructions within a single query.
This trend indicates that the simultaneous execution of multiple visual instructions is more demanding for current models than single-instruction settings, even when individual instructions are handled correctly in isolation.
The observed performance degradation highlights a gap between single-instruction competence and compositional instruction understanding, indicating an important direction for future model development.

\subsection{Validity of LMM-as-a-Judge}
\label{sec:eval_valid}

To validate the reliability of using LMM as evaluators, we  analyze the correlation between LMM-based assessments and human expert judgments.
Specifically, we randomly sample results from two representative models, Nano Banana Pro and GPT-Image-1.
For each model, we select 10 edited samples from each of the 10 tasks, resulting in 100 evaluated samples per model.
Four human experts independently assess all selected samples using the same evaluation criteria and metrics as the LMM evaluator.
The annotation interface used for human evaluation is illustrated in Figure~\ref{fig:app_human}.
Human scores are obtained by averaging the scores across the four experts.
We then compute the Pearson correlation coefficient between the human-annotated scores and the scores produced by the LMM evaluator.

As shown in Figure~\ref{fig:human}, our LMM-based evaluations exhibit strong correlation with human judgments.
The overall Pearson correlation coefficient across all samples reaches $r=0.9602$.
When analyzed by model, the correlation remains consistently high, with $r=0.9673$ for Nano Banana Pro and $r=0.9531$ for GPT-Image-1.
These results indicate a high level of agreement between the LMM evaluator and human experts across different tasks and models.
Together, this analysis demonstrates that the proposed LMM-as-a-Judge framework provides reliable and human-aligned evaluations across diverse tasks and models.

\subsection{Synergy Between Textual and Visual Instructions}

We further conduct a qualitative analysis to examine how textual and visual instructions interact in guiding image editing.
Figure~\ref{fig:instruction} presents two representative cases from state-of-the-art models that illustrate the complementary and synergistic roles of the two instruction modalities.

In the first case, we observe that providing a highly detailed textual instruction alone does not guarantee correct editing behavior.
Despite explicitly specifying the editing target in text, the model fails to localize the intended region.
In contrast, when a simple visual instruction is used to directly indicate the target object, a much shorter and less detailed textual prompt suffices to produce the correct result.
This example suggests that visual instructions provide a strong grounding signal for target localization, effectively reducing ambiguity that is difficult to resolve through language alone.

The second case reveals a different failure mode.
Here, neither a detailed textual instruction alone nor a combination of a brief textual instruction with a visual cue leads to a correct outcome.
Only when a detailed textual instruction is paired with an explicit visual instruction does the model succeed.
This indicates that visual instructions may benefit from complementary textual specification when complex semantic constraints need to be expressed.

Together, these examples highlight that \textbf{textual and visual instructions play distinct yet complementary roles in image editing.}
Visual instructions enable precise spatial grounding, while textual instructions are essential for conveying semantic intent and complementary information.
Additional qualitative cases with visually embedded instructions are provided in Appendix~\ref{sec:app_instruction}.

\section{Conclusion}

In this paper, we introduced \abbr, a benchmark designed to systematically evaluate visual instruction-following capabilities in image editing.
By organizing tasks into a three-level hierarchy consisting of the \textit{Deictic}, \textit{Morphological}, and \textit{Causal} Level, we provide a structured framework for assessing increasingly complex forms of visual–linguistic interaction.
Through extensive experiments on a wide range of state-of-the-art proprietary and open-source models, we show that current systems demonstrate early-stage competence on explicit, localized visual instructions, while performance degrades as interaction complexity increases.
Our analysis further reveals pronounced differences across models, visual styles, and task compositions, highlighting challenges in reasoning, compositional instruction execution, and multi-task coordination.
We hope that \abbr will serve as a useful testbed for future research, encouraging the development of models with improved visual instruction-following and coordination with textual guidance.

\bibliography{example_paper}

@inproceedings{ku2024viescore,
  title={Viescore: Towards explainable metrics for conditional image synthesis evaluation},
  author={Ku, Max and Jiang, Dongfu and Wei, Cong and Yue, Xiang and Chen, Wenhu},
  booktitle={Proceedings of the 62nd Annual Meeting of the Association for Computational Linguistics (Volume 1: Long Papers)},
  pages={12268--12290},
  year={2024}
}

@article{zhang2025scaling,
  title={Scaling and Beyond: Advancing Spatial Reasoning in MLLMs Requires New Recipes},
  author={Zhang, Huanyu and Li, Chengzu and Wu, Wenshan and Mao, Shaoguang and Zhang, Yifan and Tian, Haochen and Vuli{\'c}, Ivan and Zhang, Zhang and Wang, Liang and Tan, Tieniu and others},
  journal={arXiv preprint arXiv:2504.15037},
  year={2025}
}

@inproceedings{li2024mvot,
  title={Imagine while Reasoning in Space: Multimodal Visualization-of-Thought},
  author={Li, Chengzu and Wu, Wenshan and Zhang, Huanyu and Xia, Yan and Mao, Shaoguang and Dong, Li and Vuli{\'c}, Ivan and Wei, Furu},
  booktitle={The Forty-Second International Conference on Machine Learning},
  year={2025}
}

@inproceedings{zhangmme,
  title={MME-RealWorld: Could Your Multimodal LLM Challenge High-Resolution Real-World Scenarios that are Difficult for Humans?},
  author={Zhang, YiFan and Zhang, Huanyu and Tian, Haochen and Fu, Chaoyou and Zhang, Shuangqing and Wu, Junfei and Li, Feng and Wang, Kun and Wen, Qingsong and Zhang, Zhang and others},
  booktitle={The Thirteenth International Conference on Learning Representations},
  year={2025}
}

@article{zhang2025latent,
  title={Latent Sketchpad: Sketching Visual Thoughts to Elicit Multimodal Reasoning in MLLMs},
  author={Zhang, Huanyu and Wu, Wenshan and Li, Chengzu and Shang, Ning and Xia, Yan and Huang, Yangyu and Zhang, Yifan and Dong, Li and Zhang, Zhang and Wang, Liang and others},
  journal={arXiv preprint arXiv:2510.24514},
  year={2025}
}

@article{zhao2025envisioning,
  title={Envisioning beyond the pixels: Benchmarking reasoning-informed visual editing},
  author={Zhao, Xiangyu and Zhang, Peiyuan and Tang, Kexian and Zhu, Xiaorong and Li, Hao and Chai, Wenhao and Zhang, Zicheng and Xia, Renqiu and Zhai, Guangtao and Yan, Junchi and others},
  journal={arXiv preprint arXiv:2504.02826},
  year={2025}
}

@inproceedings{xiao2025omnigen,
  title={Omnigen: Unified image generation},
  author={Xiao, Shitao and Wang, Yueze and Zhou, Junjie and Yuan, Huaying and Xing, Xingrun and Yan, Ruiran and Li, Chaofan and Wang, Shuting and Huang, Tiejun and Liu, Zheng},
  booktitle={Proceedings of the Computer Vision and Pattern Recognition Conference},
  pages={13294--13304},
  year={2025}
}

@article{wu2025omnigen2,
  title={OmniGen2: Exploration to Advanced Multimodal Generation},
  author={Wu, Chenyuan and Zheng, Pengfei and Yan, Ruiran and Xiao, Shitao and Luo, Xin and Wang, Yueze and Li, Wanli and Jiang, Xiyan and Liu, Yexin and Zhou, Junjie and others},
  journal={arXiv preprint arXiv:2506.18871},
  year={2025}
}

@article{lin2025uniworld,
  title={Uniworld: High-resolution semantic encoders for unified visual understanding and generation},
  author={Lin, Bin and Li, Zongjian and Cheng, Xinhua and Niu, Yuwei and Ye, Yang and He, Xianyi and Yuan, Shenghai and Yu, Wangbo and Wang, Shaodong and Ge, Yunyang and others},
  journal={arXiv preprint arXiv:2506.03147},
  year={2025}
}

@article{deng2025bagel,
  title={Emerging properties in unified multimodal pretraining},
  author={Deng, Chaorui and Zhu, Deyao and Li, Kunchang and Gou, Chenhui and Li, Feng and Wang, Zeyu and Zhong, Shu and Yu, Weihao and Nie, Xiaonan and Song, Ziang and others},
  journal={arXiv preprint arXiv:2505.14683},
  year={2025}
}

@article{li2025editr1,
  title={Uniworld-v2: Reinforce image editing with diffusion negative-aware finetuning and mllm implicit feedback},
  author={Li, Zongjian and Liu, Zheyuan and Zhang, Qihui and Lin, Bin and Wu, Feize and Yuan, Shenghai and Yan, Zhiyuan and Ye, Yang and Yu, Wangbo and Niu, Yuwei and others},
  journal={arXiv preprint arXiv:2510.16888},
  year={2025}
}

@article{wu2025qwen,
  title={Qwen-image technical report},
  author={Wu, Chenfei and Li, Jiahao and Zhou, Jingren and Lin, Junyang and Gao, Kaiyuan and Yan, Kun and Yin, Sheng-ming and Bai, Shuai and Xu, Xiao and Chen, Yilei and others},
  journal={arXiv preprint arXiv:2508.02324},
  year={2025}
}

@misc{blackforestlabs2025flux2,
  title = {FLUX.2: Frontier Visual Intelligence},
  author = {{Black Forest Labs}},
  year = {2025},
  month = {December},
  howpublished = {\url{https://bfl.ai/blog/flux-2}},
}

@misc{banana,
  title = {Introducing gemini 2.5 flash image, our state
of-the-art image model.},
  author = {{Google}},
  year = {2025},
  month = {August},
  howpublished = {\url{https://developers.googleblog.com/introducing-gemini-2-5-flash-image/}},
}

@misc{bananapro,
  title = {Introducing nano banana pro},
  author = {{Google}},
  year = {2025},
  month = {November},
  howpublished = {\url{https://blog.google/technology/ai/nano-banana-pro/}},
}

@misc{gemini,
  title = {A new era of intelligence with Gemini 3},
  author = {{Google}},
  year = {2025},
  month = {November},
  howpublished = {\url{https://blog.google/products-and-platforms/products/gemini/gemini-3/}},
}

@misc{GPT5,
  title = {GPT5 is here},
  author = {{OpenAI}},
  year = {2025},
  month = {August},
  howpublished = {\url{https://openai.com/gpt-5/}},
}

@article{fang2025got,
  title={Got: Unleashing reasoning capability of multimodal large language model for visual generation and editing},
  author={Fang, Rongyao and Duan, Chengqi and Wang, Kun and Huang, Linjiang and Li, Hao and Yan, Shilin and Tian, Hao and Zeng, Xingyu and Zhao, Rui and Dai, Jifeng and others},
  journal={arXiv preprint arXiv:2503.10639},
  year={2025}
}

@inproceedings{
feng2025videor,
title={Video-R1: Reinforcing Video Reasoning in {MLLM}s},
author={Kaituo Feng and Kaixiong Gong and Bohao Li and Zonghao Guo and Yibing Wang and Tianshuo Peng and Junfei Wu and Xiaoying Zhang and Benyou Wang and Xiangyu Yue},
booktitle={The Thirty-ninth Annual Conference on Neural Information Processing Systems},
year={2025},
url={https://openreview.net/forum?id=a2JTVVvcEl}
}

@inproceedings{
wu2025reinforcing,
title={Reinforcing Spatial Reasoning in Vision-Language Models with Interwoven Thinking and Visual Drawing},
author={Junfei Wu and Jian Guan and Kaituo Feng and Qiang Liu and Shu Wu and Liang Wang and Wei Wu and Tieniu Tan},
booktitle={The Thirty-ninth Annual Conference on Neural Information Processing Systems},
year={2025},
url={https://openreview.net/forum?id=yyWeSAsOhs}
}

@article{chen2025opengpt,
  title={Opengpt-4o-image: A comprehensive dataset for advanced image generation and editing},
  author={Chen, Zhihong and Bai, Xuehai and Shi, Yang and Fu, Chaoyou and Zhang, Huanyu and Wang, Haotian and Sun, Xiaoyan and Zhang, Zhang and Wang, Liang and Zhang, Yuanxing and others},
  journal={arXiv preprint arXiv:2509.24900},
  year={2025}
}

@article{ye2025imgedit,
  title={Imgedit: A unified image editing dataset and benchmark},
  author={Ye, Yang and He, Xianyi and Li, Zongjian and Lin, Bin and Yuan, Shenghai and Yan, Zhiyuan and Hou, Bohan and Yuan, Li},
  journal={arXiv preprint arXiv:2505.20275},
  year={2025}
}

@article{zhang2025mmrlhf,
  title={Mm-rlhf: The next step forward in multimodal llm alignment},
  author={Zhang, Yi-Fan and Yu, Tao and Tian, Haochen and Fu, Chaoyou and Li, Peiyan and Zeng, Jianshu and Xie, Wulin and Shi, Yang and Zhang, Huanyu and Wu, Junkang and others},
  journal={arXiv preprint arXiv:2502.10391},
  year={2025}
}

@article{seedream2025seedream,
  title={Seedream 4.0: Toward next-generation multimodal image generation},
  author={Seedream, Team and Chen, Yunpeng and Gao, Yu and Gong, Lixue and Guo, Meng and Guo, Qiushan and Guo, Zhiyao and Hou, Xiaoxia and Huang, Weilin and Huang, Yixuan and others},
  journal={arXiv preprint arXiv:2509.20427},
  year={2025}
}

@article{hurst2024gpt,
  title={Gpt-4o system card},
  author={Hurst, Aaron and Lerer, Adam and Goucher, Adam P and Perelman, Adam and Ramesh, Aditya and Clark, Aidan and Ostrow, AJ and Welihinda, Akila and Hayes, Alan and Radford, Alec and others},
  journal={arXiv preprint arXiv:2410.21276},
  year={2024}
}

@misc{wan,
  title = {Wan Image Edit},
  author = {{Wan}},
  year = {2025},
  month = {November},
  howpublished = {\url{https://wan.video/}},
}

@inproceedings{sheynin2024emu,
  title={Emu edit: Precise image editing via recognition and generation tasks},
  author={Sheynin, Shelly and Polyak, Adam and Singer, Uriel and Kirstain, Yuval and Zohar, Amit and Ashual, Oron and Parikh, Devi and Taigman, Yaniv},
  booktitle={Proceedings of the IEEE/CVF Conference on Computer Vision and Pattern Recognition},
  pages={8871--8879},
  year={2024}
}

@article{zhang2025mm,
  title={Mm-rlhf: The next step forward in multimodal llm alignment},
  author={Zhang, Yi-Fan and Yu, Tao and Tian, Haochen and Fu, Chaoyou and Li, Peiyan and Zeng, Jianshu and Xie, Wulin and Shi, Yang and Zhang, Huanyu and Wu, Junkang and others},
  journal={arXiv preprint arXiv:2502.10391},
  year={2025}
}

@article{zhang2023magicbrush,
  title={Magicbrush: A manually annotated dataset for instruction-guided image editing},
  author={Zhang, Kai and Mo, Lingbo and Chen, Wenhu and Sun, Huan and Su, Yu},
  journal={Advances in Neural Information Processing Systems},
  volume={36},
  pages={31428--31449},
  year={2023}
}

@article{wu2024visual,
  title={Visual prompting in multimodal large language models: A survey},
  author={Wu, Junda and Zhang, Zhehao and Xia, Yu and Li, Xintong and Xia, Zhaoyang and Chang, Aaron and Yu, Tong and Kim, Sungchul and Rossi, Ryan A and Zhang, Ruiyi and others},
  journal={arXiv preprint arXiv:2409.15310},
  year={2024}
}

@article{zhang2023vpgtrans,
  title={Vpgtrans: Transfer visual prompt generator across llms},
  author={Zhang, Ao and Fei, Hao and Yao, Yuan and Ji, Wei and Li, Li and Liu, Zhiyuan and Chua, Tat-Seng},
  journal={Advances in Neural Information Processing Systems},
  volume={36},
  pages={20299--20319},
  year={2023}
}

@inproceedings{wu2024dettoolchain,
  title={Dettoolchain: A new prompting paradigm to unleash detection ability of mllm},
  author={Wu, Yixuan and Wang, Yizhou and Tang, Shixiang and Wu, Wenhao and He, Tong and Ouyang, Wanli and Torr, Philip and Wu, Jian},
  booktitle={European Conference on Computer Vision},
  pages={164--182},
  year={2024},
  organization={Springer}
}

@article{jiang2024joint,
  title={Joint visual and text prompting for improved object-centric perception with multimodal large language models},
  author={Jiang, Songtao and Zhang, Yan and Zhou, Chenyi and Jin, Yeying and Feng, Yang and Wu, Jian and Liu, Zuozhu},
  journal={arXiv preprint arXiv:2404.04514},
  year={2024}
}

@article{zhou2024minedreamer,
  title={Minedreamer: Learning to follow instructions via chain-of-imagination for simulated-world control},
  author={Zhou, Enshen and Qin, Yiran and Yin, Zhenfei and Huang, Yuzhou and Zhang, Ruimao and Sheng, Lu and Qiao, Yu and Shao, Jing},
  journal={arXiv preprint arXiv:2403.12037},
  year={2024}
}

@article{huang2025diffusion,
  title={Diffusion model-based image editing: A survey},
  author={Huang, Yi and Huang, Jiancheng and Liu, Yifan and Yan, Mingfu and Lv, Jiaxi and Liu, Jianzhuang and Xiong, Wei and Zhang, He and Cao, Liangliang and Chen, Shifeng},
  journal={IEEE Transactions on Pattern Analysis and Machine Intelligence},
  year={2025},
  publisher={IEEE}
}

@article{shuai2024survey,
  title={A survey of multimodal-guided image editing with text-to-image diffusion models},
  author={Shuai, Xincheng and Ding, Henghui and Ma, Xingjun and Tu, Rongcheng and Jiang, Yu-Gang and Tao, Dacheng},
  journal={arXiv preprint arXiv:2406.14555},
  year={2024}
}

@article{zhao2025risebench,
  title={Envisioning beyond the pixels: Benchmarking reasoning-informed visual editing},
  author={Zhao, Xiangyu and Zhang, Peiyuan and Tang, Kexian and Zhu, Xiaorong and Li, Hao and Chai, Wenhao and Zhang, Zicheng and Xia, Renqiu and Zhai, Guangtao and Yan, Junchi and others},
  journal={arXiv preprint arXiv:2504.02826},
  year={2025}
}

@article{wu2025kris,
  title={KRIS-Bench: Benchmarking Next-Level Intelligent Image Editing Models},
  author={Wu, Yongliang and Li, Zonghui and Hu, Xinting and Ye, Xinyu and Zeng, Xianfang and Yu, Gang and Zhu, Wenbo and Schiele, Bernt and Yang, Ming-Hsuan and Yang, Xu},
  journal={arXiv preprint arXiv:2505.16707},
  year={2025}
}

@book{buxton2010sketching,
  title={Sketching user experiences: getting the design right and the right design},
  author={Buxton, Bill},
  year={2010},
  publisher={Morgan kaufmann}
}

@article{li202511plus,
  title={11plus-bench: Demystifying multimodal llm spatial reasoning with cognitive-inspired analysis},
  author={Li, Chengzu and Wu, Wenshan and Zhang, Huanyu and Li, Qingtao and Gao, Zeyu and Xia, Yan and Hern{\'a}ndez-Orallo, Jos{\'e} and Vuli{\'c}, Ivan and Wei, Furu},
  journal={arXiv preprint arXiv:2508.20068},
  year={2025}
}

@inproceedings{gu2019mask,
  title={Mask-guided portrait editing with conditional gans},
  author={Gu, Shuyang and Bao, Jianmin and Yang, Hao and Chen, Dong and Wen, Fang and Yuan, Lu},
  booktitle={Proceedings of the IEEE/CVF conference on computer vision and pattern recognition},
  pages={3436--3445},
  year={2019}
}

@incollection{tversky2013visualizing,
  title={Visualizing thought},
  author={Tversky, Barbara},
  booktitle={Handbook of human centric visualization},
  pages={3--40},
  year={2013},
  publisher={Springer}
}

@incollection{herring2015new,
  title={New frontiers in interactive multimodal communication},
  author={Herring, Susan C},
  booktitle={The Routledge handbook of language and digital communication},
  pages={398--402},
  year={2015},
  publisher={Routledge}
}

@article{liu2025step1x,
  title={Step1x-edit: A practical framework for general image editing},
  author={Liu, Shiyu and Han, Yucheng and Xing, Peng and Yin, Fukun and Wang, Rui and Cheng, Wei and Liao, Jiaqi and Wang, Yingming and Fu, Honghao and Han, Chunrui and others},
  journal={arXiv preprint arXiv:2504.17761},
  year={2025}
}

@misc{li2026thinkingframesvisualcontext,
      title={Thinking in Frames: How Visual Context and Test-Time Scaling Empower Video Reasoning}, 
      author={Chengzu Li and Zanyi Wang and Jiaang Li and Yi Xu and Han Zhou and Huanyu Zhang and Ruichuan An and Dengyang Jiang and Zhaochong An and Ivan Vulić and Serge Belongie and Anna Korhonen},
      year={2026},
      eprint={2601.21037},
      archivePrefix={arXiv},
      primaryClass={cs.LG},
      url={https://arxiv.org/abs/2601.21037}, 
}

@inproceedings{li-etal-2024-topviewrs,
    title = "{T}op{V}iew{RS}: Vision-Language Models as Top-View Spatial Reasoners",
    author = "Li, Chengzu  and
      Zhang, Caiqi  and
      Zhou, Han  and
      Collier, Nigel  and
      Korhonen, Anna  and
      Vuli{\'c}, Ivan",
    editor = "Al-Onaizan, Yaser  and
      Bansal, Mohit  and
      Chen, Yun-Nung",
    booktitle = "Proceedings of the 2024 Conference on Empirical Methods in Natural Language Processing",
    month = nov,
    year = "2024",
    address = "Miami, Florida, USA",
    publisher = "Association for Computational Linguistics",
    url = "https://aclanthology.org/2024.emnlp-main.106/",
    doi = "10.18653/v1/2024.emnlp-main.106",
    pages = "1786--1807"
}
\bibliographystyle{icml2026}

\newpage
\appendix
\onecolumn

\section{Related Work}

\sparagraph{Image Editing  with Generative Models} Building on deep generative models, the image editing literature has explored how to modify existing images in a controllable and semantic manner. Early approaches leveraged conditional GANs for localized editing, such as mask-guided portrait manipulation, demonstrating how learned generative priors can support targeted changes~\cite{gu2019mask}. Recent methods increasingly utilize diffusion-based frameworks for editing tasks, leveraging the iterative denoising process to incorporate multimodal guidance (e.g., text or exemplar images) and achieve high-quality edits while preserving source content~\cite{huang2025diffusion}.
While traditional generator architectures are optimized for individual tasks (e.g., text-to-image or specific editing objectives), there has been growing interest in unified models that integrate multiple generative and editing capabilities within a single framework~\cite{bananapro,wan,seedream2025seedream}. 
These unified approaches represent an emerging direction toward general-purpose visual generative and transformative models that seamlessly integrate content creation and editing.

\sparagraph{Image Editing Benchmarks} Most existing image editing benchmarks rely exclusively on textual prompts as editing instructions and do not explicitly model visual prompts~\cite{huang2025diffusion,shuai2024survey}. Consequently, they are limited in evaluating advanced models’ ability to understand and follow visual prompts for image editing. Early benchmarks focus on a small set of primitive editing operations and limited task diversity. Representative examples include MagicBrush~\cite{zhang2023magicbrush} and EMU-Edit~\cite{sheynin2024emu}, which primarily assess models’ performance on basic add, remove, and replace operations, and thus provide only a coarse evaluation of editing capabilities. More recent benchmarks expand the scope of editing tasks and improve evaluation protocols. ImgEdit-Bench~\cite{ye2025imgedit} and GEdit-Bench~\cite{liu2025step1x} extend the range of editing types and adopt a VLM-as-a-judge paradigm to better capture semantic correctness. Beyond task diversity, several benchmarks are designed to assess reasoning grounded in world knowledge. RISEBench~\cite{zhao2025risebench} evaluates temporal, spatial, and causal editing capabilities, while KRISBench~\cite{wu2025kris} introduces a knowledge-based taxonomy that covers conceptual, factual, and procedural editing types.


\sparagraph{Multimodal Interaction}
Multimodal interaction lies at the core of contemporary research on models that jointly reason about language and vision.
With the emergence of large multimodal models (MLLMs)~\cite{GPT5,gemini,zhang2025mmrlhf}, interaction paradigms have expanded toward richer and more integrated reasoning over text and images~\cite{zhangmme,feng2025videor,zhang2025scaling,zhang2025mm}. 
In particular, recent frameworks such as Thinking with Images propose that models should not only “see” images as static inputs, but incorporate visual information as intermediate steps in their reasoning process~\cite{zhang2025latent,li2024mvot,wu2025reinforcing,zhou2024minedreamer,li2026thinkingframesvisualcontext}. 
This perspective has been articulated as a new paradigm where models leverage visual representations dynamically within multi-step reasoning processes to better handle complex tasks that cannot be solved through text alone. 
Complementary to this, recent works in multimodal reasoning and generation explore how multimodal chain-of-thought (CoT) and reasoning planning can be explicitly structured to improve performance. 
For example, GoT (Generation Chain-of-Thought)~\cite{fang2025got} introduces a visual reasoning pipeline where a Multimodal LLM generates structured intermediate reasoning steps before synthesis or editing, enabling fine-grained semantic and spatial control.
While recent advances in generative and editing models have made significant progress in responding to textual instructions, existing work predominantly focuses on optimizing text-driven generation and image editing~\cite{wu2025qwen,seedream2025seedream,chen2025opengpt}.
This gap is particularly pronounced in spatially grounded editing scenarios, where explicit visual cues such as bounding boxes, arrows, or sketches carry essential information that cannot be fully captured by text alone~\cite{wu2024visual,jiang2024joint,zhang2023vpgtrans,wu2024dettoolchain,li202511plus}.
To address this gap, we propose \abbr, the first benchmark specifically designed to systematically evaluate visual instruction-guided image editing.

\section{Data Collection and Annotation}
\label{app:data}
\subsection{Deictic Level}

The four Deictic Level tasks (Addition, Removal, Replacement, and Translation) share the same set of 100 source images.
All source images are collected from publicly available online resources and subsequently curated through manual filtering to ensure visual clarity and suitability for localized editing.
The final dataset consists of 34 real-world images, 33 animation-style images, and 33 sketch-style images, providing a balanced coverage of diverse visual styles.

All visual instructions in the Deictic Level are annotated manually.
To minimize potential bias introduced by color variation and to ensure consistency across tasks, all annotations are rendered exclusively in red.
Different annotation primitives are used depending on the task type.
For the Addition task, annotators draw a bounding box at the target location where new visual content should be introduced.
For the Removal and Replacement tasks, annotators draw a bounding box tightly enclosing the object or region to be removed or replaced.
For the Translation task, annotators first draw a bounding box around the target object, and then add an arrow indicating the desired destination to which the object should be relocated.
In all cases, annotations are designed to be spatially explicit and unambiguous, providing clear grounding cues for the intended edit.

\subsection{Morphological Level}

\paragraph{Pose Control.}
For the Pose Control task, we construct source-reference image pairs to explicitly evaluate pose transfer while preserving character identity.
We first collect 27 images of human subjects from publicly available online sources, ensuring that each image contains a single, clearly visible person with minimal occlusion.
In addition, we collect 26 distinct pose reference images in the form of schematic, stick-figure–like pose diagrams, each depicting a unique body configuration.
Source images and pose references are then manually paired to form 100 source–reference image pairs.
This manual pairing process ensures sufficient pose diversity while avoiding trivial or ambiguous pose correspondences.

\paragraph{Reorientation.}
For the Reorientation task, we collect images of objects with clearly defined facing directions.
These images include, but are not limited to, vehicles, chairs, cameras, humans, and shoes, where orientation can be unambiguously inferred from semantic cues.
For each image, we manually annotate a viewing frustum indicating the target orientation.
The annotated frustums are carefully designed to be visually clear and semantically meaningful, specifying yaw, pitch, and roll directions where applicable.
This process results in 100 distinct reorientation cases.

\paragraph{Draft Instantiation.}
For the Draft Instantiation task, we collect 100 source images spanning diverse visual styles, including 40 real-world images, 34 animation-style images, and 26 sketch-style images.
For each image, annotators manually draw draft-level visual instructions directly on top of the source image to indicate the desired structural modifications.
These drafts are intentionally sparse and schematic, serving as abstract blueprints rather than detailed renderings.
All draft annotations are created manually to ensure faithful representation of the intended structural guidance and consistency across styles.

\subsection{Causal Level}

\paragraph{Light Control.}
For the Light Control task, we collect images from publicly available online sources that exhibit a single, clearly identifiable light source and a well-defined subject whose appearance reflects the lighting direction.
Selected scenes are required to contain visible shading, shadows, or highlights that can meaningfully convey illumination changes.
For each image, annotators manually draw an arrow to indicate the target lighting direction, representing the propagation direction from the light source toward the subject.
The annotated lighting directions are deliberately chosen to differ from the original scene lighting while remaining physically plausible and visually interpretable.
This process results in 100 annotated samples for the Light Control task.

\paragraph{Flow Simulation.}
For the Flow Simulation task, we collect images containing objects that are sensitive to wind effects, such as human hair, hanging clothes, dandelions, and burning candles.
These objects provide clear visual cues for inferring airflow direction through deformation or motion.
Similar to Light Control, all visual instructions are manually annotated by drawing arrows that indicate the target wind direction.
Annotations are designed to be distinct from the original scene conditions and to induce meaningful, causally consistent visual changes.
In total, 100 annotated samples are constructed for the Flow Simulation task.

\paragraph{Billiards.}
For the Billiards task, we generate data through a procedural simulation pipeline.
Each scene consists of a fixed rectangular environment containing one white cue ball and five gray balls with distinct numeric labels.
A directional force is applied to the white ball, which is visualized by an arrow in the input image.
The white ball subsequently undergoes multiple reflections against the scene boundaries before colliding with one of the gray balls.

We control the number of boundary reflections to range from two to seven in order to vary the difficulty of the task.
The complete trajectory of the white ball is rendered as a green dashed line, and the final collision target is indicated by a red bounding box around the impacted gray ball, forming the corresponding label image.
All generated samples are manually inspected to ensure clarity of the trajectory and collision outcome.
Cases with excessive overlap or visually ambiguous trajectories are filtered out.
After this quality control process, the final Billiards dataset contains 134 distinct cases.

\section{Additional Experiments and Analysis}
\label{app:exp}

\begin{table*}[t]
\caption{ \textbf{Experimental results on \abbr.} We report task-wise and overall performance as mean $\pm$ standard deviation over 3 independent runs across the \textit{Deictic}, \textit{Morphological}, and \textit{Causal} levels. 
}
\vspace{-2mm}
\label{tab:main_app}

\rowcolors{4}{white}{gray!15}
\renewcommand{\arraystretch}{1.2}
\newcolumntype{C}{>{\centering\arraybackslash}p{0.08\textwidth}}
\newcolumntype{E}{>{\centering\arraybackslash}p{0.07\textwidth}}
\resizebox{\linewidth}{!}{
\begin{tabular}{l|cccc|ccc|ccc} 
\hline
\multirow{2}{*}{Model}       & \multicolumn{4}{c|}{Deictic Level}                             & \multicolumn{3}{c|}{Morphological Level}                & \multicolumn{3}{c}{Causal Level}                  \\
                             & AD         & RM         & RP         & TR         & PC         & RO         & DI         & LC         & FS         & BI          \\ 
\hline
Nano Banana Pro              & 82.17±0.66 & 94.07±0.73 & 88.26±1.97 & 74.80±2.13 & 72.33±0.32 & 36.04±2.58 & 88.02±2.08 & 60.34±4.78 & 59.25±3.62 & 15.92±0.78  \\
Nano Banana                  & 81.34±0.91 & 93.50±0.71 & 79.05±1.43 & 46.53±1.93 & 67.71±1.03 & 33.45±1.43 & 85.60±0.75 & 34.75±4.96 & 52.64±0.87 & 1.87±0.99   \\
GPT-image-1                  & 55.61±2.11 & 69.00±1.00 & 62.63±2.82 & 47.00±4.62 & 64.39±2.74 & 11.09±2.51 & 77.32±2.36 & 25.18±6.71 & 39.48±6.07 & 4.73±0.43   \\
Seedream 4.5                 & 81.24±0.50 & 95.82±0.75 & 81.82±3.36 & 48.93±1.18 & 66.79±2.38 & 20.11±5.62 & 82.33±2.17 & 50.50±2.60 & 45.55±1.07 & 2.99±0.00   \\
Seedream 4.0                 & 74.02±1.39 & 93.04±0.96 & 79.35±1.82 & 33.29±4.28 & 58.27±2.27 & 30.37±4.47 & 72.09±2.11 & 47.00±3.46 & 43.59±3.18 & 4.11±0.65   \\
Wan2.6                       & 66.01±3.10 & 92.90±0.85 & 68.15±0.83 & 40.95±2.69 & 59.66±2.46 & 34.89±4.54 & 80.23±2.63 & 44.46±3.45 & 50.79±0.84 & 9.08±1.20   \\
Wan2.5                       & 73.59±2.47 & 96.90±1.16 & 76.99±1.60 & 36.80±1.41 & 55.76±3.44 & 25.77±1.55 & 78.78±0.48 & 33.33±4.63 & 51.98±0.37 & 7.84±0.37   \\ 
\hline
FLUX2-dev                    & 64.57±1.18 & 8.00±1.73  & 54.4±1.71  & 5.58±1.09  & 28.76±3.20 & 22.77±1.15 & 60.68±1.89 & 33.74±3.65 & 30.81±2.54 & 2.36±0.43   \\
Qwen-Image-Edit-2509         & 55.28±2.40 & 14.38±3.05 & 30.13±0.92 & 14.48±0.57 & 15.14±1.21 & 17.67±4.29 & 21.38±1.77 & 28.00±1.32 & 44.40±1.83 & 2.36±0.78   \\
Qwen-Image-Edit              & 44.20±2.69 & 24.88±5.24 & 30.48±2.33 & 11.11±2.41 & -          & 21.33±2.17 & 54.65±0.96 & 22.00±1.80 & 32.38±1.41 & 3.11±0.57   \\
Edit-R1-Qwen-Image-Edit-2509 & 56.77±4.87 & 4.86±1.79  & 29.47±2.11 & 11.33±3.69 & 16.23±0.79 & 20.27±3.99 & 15.42±0.79 & 25.67±0.76 & 40.22±1.59 & 2.24±0.37   \\
BAGEL-think                  & 40.44±2.13 & 14.59±0.74 & 35.38±2.39 & 14.46±2.61 & 8.50±2.62  & 23.6±3.85  & 50.34±0.50 & 21.17±2.36 & 28.04±1.57 & 5.22±0.37   \\
BAGEL                        & 33.87±2.75 & 11.33±1.73 & 29.26±1.92 & 14.05±1.71 & 7.61±1.32  & 28.05±5.06 & 48.23±1.67 & 21.57±1.41 & 33.63±2.40 & 5.97±0.65   \\
Step1X-Edit-v1p2             & 33.92±0.35 & 12.59±2.18 & 28.17±1.29 & 13.52±2.94 & -          & 25.17±0.84 & 71.48±0.07 & 25.00±2.78 & 29.67±2.07 & 0.37±0.00   \\
OmniGen2                     & 26.29±1.07 & 26.20±2.44 & 20.84±0.58 & 4.51±0.98  & 11.40±1.11 & 17.44±1.69 & 30.51±0.36 & 17.33±2.02 & 17.61±5.07 & 2.74±0.77   \\
UniWorld-V1                  & 15.18±0.95 & 14.52±2.68 & 22.03±1.92 & 3.59±0.92  & 11.95±0.90 & 16.57±6.38 & 34.43±3.00 & 15.5±2.29  & 9.28±2.68  & 0.00±0.00   \\
OmniGen                      & 2.63±0.47  & 7.48±1.35  & 5.26±0.16  & 1.23±0.07  & 5.79±1.47  & 13.88±2.15 & 3.93±0.41  & 2.33±0.76  & 2.00±2.00  & 0.00±0.00   \\
\hline
\end{tabular}}
\vspace{-4mm}
\end{table*}

\subsection{Full Experimental Results}
The quantitative comparison results on the \abbr are summarized in Table~\ref{tab:main_app}. To ensure a fair and robust evaluation, we report the results as the Mean $\pm$ Standard Deviation (Mean $\pm$ SD) over 3 independent runs.

\begin{table}[t]
\caption{Complete quantitative results of style-wise performance on Draft Instantiation across proprietary models}
\label{tab:app_style_di1}
\centering
\resizebox{0.8\linewidth}{!}{
\begin{tabular}{l|ccccccccccccccccc}
\hline
           & Nano Banana Pro & Nano Banana & GPT-image-1 & Seedream 4.5 & Seedream 4.0 & Wan2.6 & Wan2.5 \\\hline
Animation  & 94.61           & 90.06       & 79.89       & 93.11        & 90.13        & 87.82  & 84.28  \\
Sketch     & 90.75           & 81.08       & 71.02       & 81.02        & 65.99        & 75.45  & 74.61  \\
Real-world & 80.34           & 84.81       & 79.40       & 73.59        & 59.81        & 76.89  & 76.55 \\ \hline
\end{tabular}
}
\end{table}

\begin{table}[t]
\caption{Complete quantitative results of style-wise performance on Draft Instantiation across open-source models}
\label{tab:app_style_di2}
\renewcommand\arraystretch{1.3}
\resizebox{\linewidth}{!}{
\begin{tabular}{l|cccccccccc}
\hline
           & FLUX2-dev & Qwen-Image-Edit-2509 & Qwen-Image-Edit & Edit-R1-Qwen-Image-Edit-2509 & BAGEL-think & BAGEL & Step1X-Edit-v1p2 & OmniGen2 & UniWorld-V1 & OmniGen \\\hline
Animation  & 73.90     & 18.97                & 65.52           & 8.31                         & 52.92       & 54.28 & 78.44            & 42.50    & 30.54       & 0.00    \\
Sketch     & 57.93     & 10.46                & 45.38           & 9.39                         & 55.32       & 48.65 & 66.70            & 5.85     & 25.39       & 2.67    \\
Real-world & 51.24     & 30.52                & 51.43           & 25.38                        & 44.91       & 42.82 & 68.67            & 36.36    & 44.93       & 8.09   \\\hline
\end{tabular}
}
\end{table}

\subsection{Style-wise Performance Analysis}
\label{app:di_style}

\begin{figure}
    \centering
    \includegraphics[width=\linewidth]{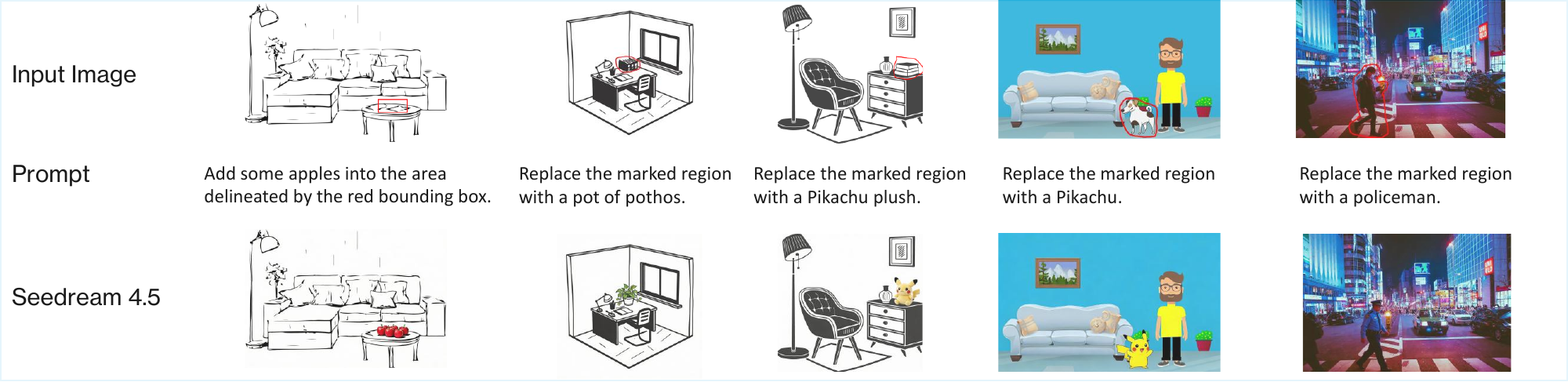}
\caption{\textbf{Examples of editing results from Seedream 4.5 across different image styles}}
    \label{fig:seed_style}
\end{figure}
We report the complete style-wise performance results of all evaluated models on the Draft Instantiation task in Tables~\ref{tab:app_style_di1} and~\ref{tab:app_style_di2}.
We further provide a qualitative analysis of style-dependent behavior on the Draft Instantiation task, with a particular focus on Seedream~4.5.
As shown in Figure~\ref{fig:seed_style}, Seedream~4.5 exhibits noticeable difficulty in preserving the original sketch style after applying the draft-based visual instructions.
Specifically, the edited regions often deviate from the input sketch domain, producing outputs with inconsistent rendering styles or mixed visual characteristics.

\begin{figure}
    \centering
    \includegraphics[width=\linewidth]{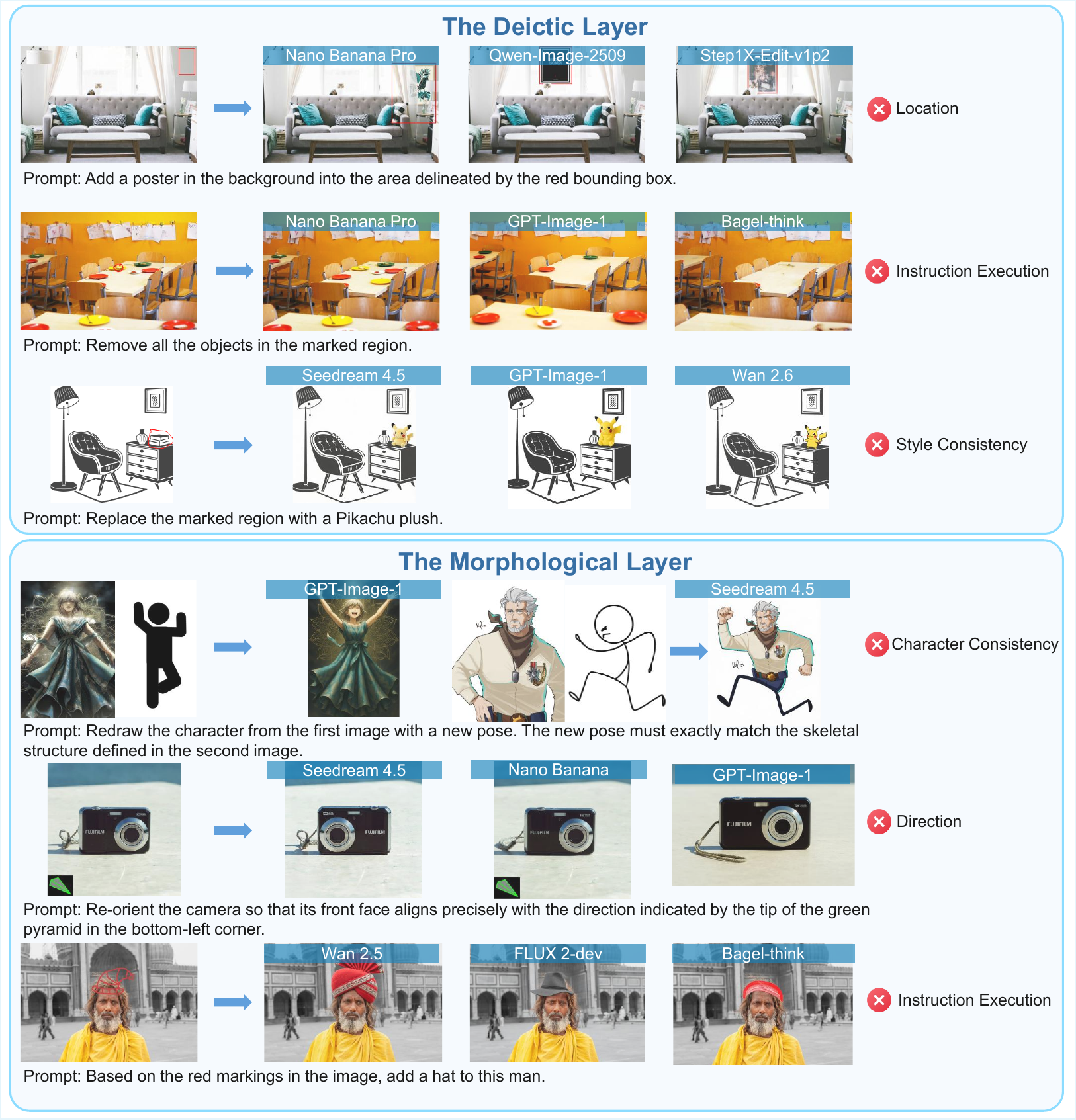}
\caption{\textbf{Qualitative incorrect examples on the Deictic and Morphological Level}}
    \label{fig:app_error1}
\end{figure}
\begin{figure}[t]
    \centering
    \includegraphics[width=\linewidth]{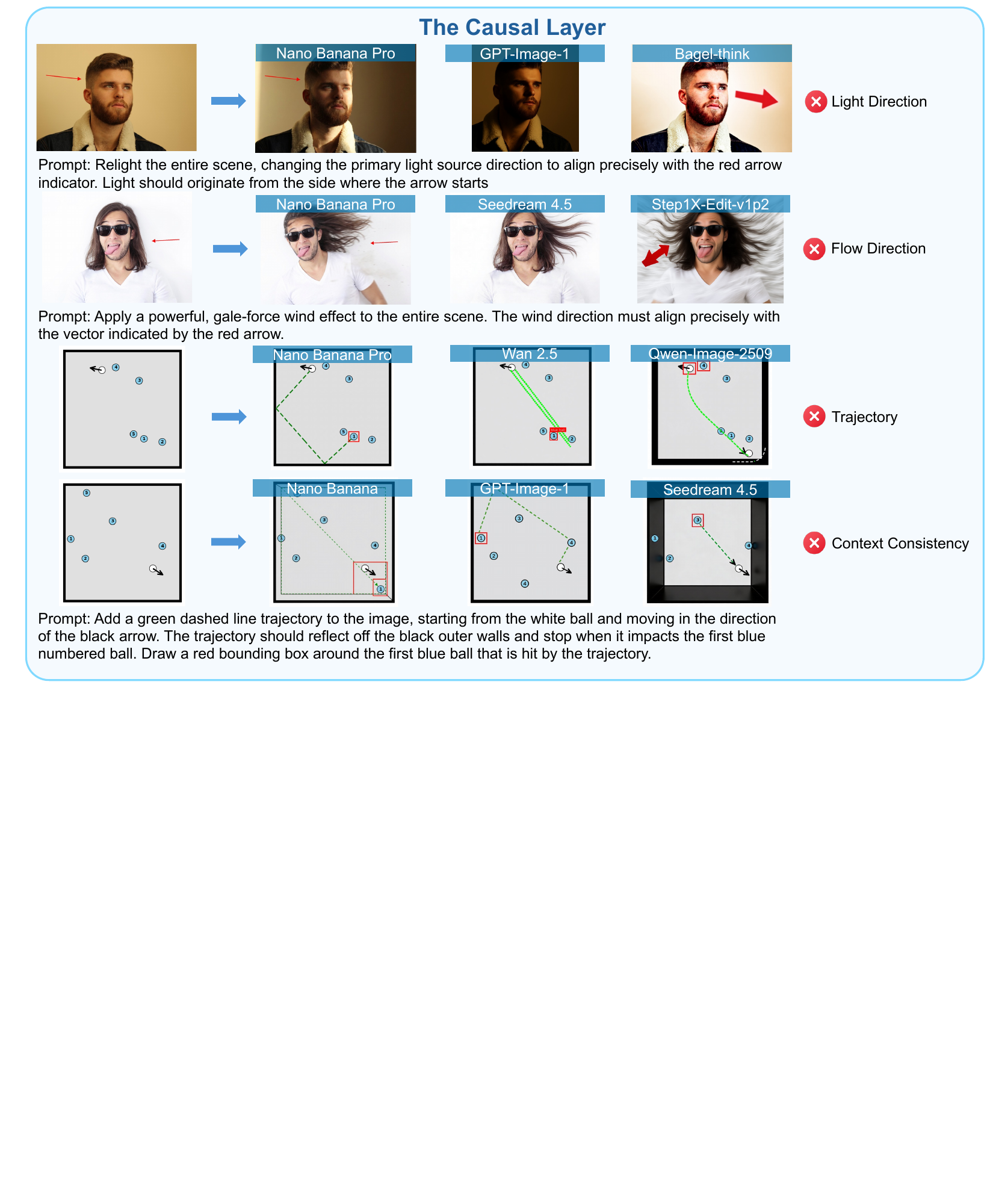}
\caption{\textbf{Qualitative incorrect examples on the Causal Level}}
    \label{fig:app_error2}
\end{figure}

\subsection{Error Analysis}

We conduct an error analysis to better understand the failure modes exhibited by current models across different interaction levels and tasks.
Representative examples are shown in Figures~\ref{fig:app_error1} and~\ref{fig:app_error2}.

\paragraph{Deictic Level.}
For the Addition task, a common failure mode is inaccurate spatial localization.
As illustrated in Figure~\ref{fig:app_error1}, models sometimes place the added object partially or entirely outside the annotated bounding box, resulting in incorrect placement.
In the Removal task, models may remove unintended content beyond the specified region, occasionally deleting objects that are not marked for editing.
Another frequent issue in both Removal and Replacement tasks is stylistic inconsistency.
In particular, for Replacement, the newly generated content may not match the visual style of the original image, leading to perceptually incoherent results, as shown in Figure~\ref{fig:app_error1}.

\paragraph{Morphological Level.}
For Pose Control, models may fail to preserve character identity, generating outputs in which the edited subject is no longer consistent with the original character.
In the Reorientation task, models sometimes struggle to align the object orientation with the annotated viewing frustum, producing results that only partially reflect the intended yaw, pitch, or roll.
For Draft Instantiation, failures often arise when the generated output does not faithfully realize the entity or structure specified by the draft-based visual instruction, resulting in incomplete or incorrect instantiations.

\paragraph{Causal Level.}
In the Light Control task, models may modify the lighting conditions of the scene but fail to align the illumination direction with the annotated arrow, leading to directionally inconsistent shading or shadows.
Similar issues are observed in Flow Simulation, where wind effects are present but do not follow the specified direction, as illustrated in Figure~\ref{fig:app_error2}.
For the Billiards task, models may predict incorrect motion trajectories, including wrong reflection sequences or target collisions.
In some cases, additional errors occur when the background or static elements of the scene are unintentionally altered, violating contextual preservation.

Overall, these error cases highlight persistent challenges in precise spatial grounding, stylistic consistency, and causal reasoning under visual instruction guidance.

\subsection{Qualitative Case Study with Visually Embedded Instructions}
\label{sec:app_instruction}

\begin{figure}
    \centering
    \includegraphics[width=\linewidth]{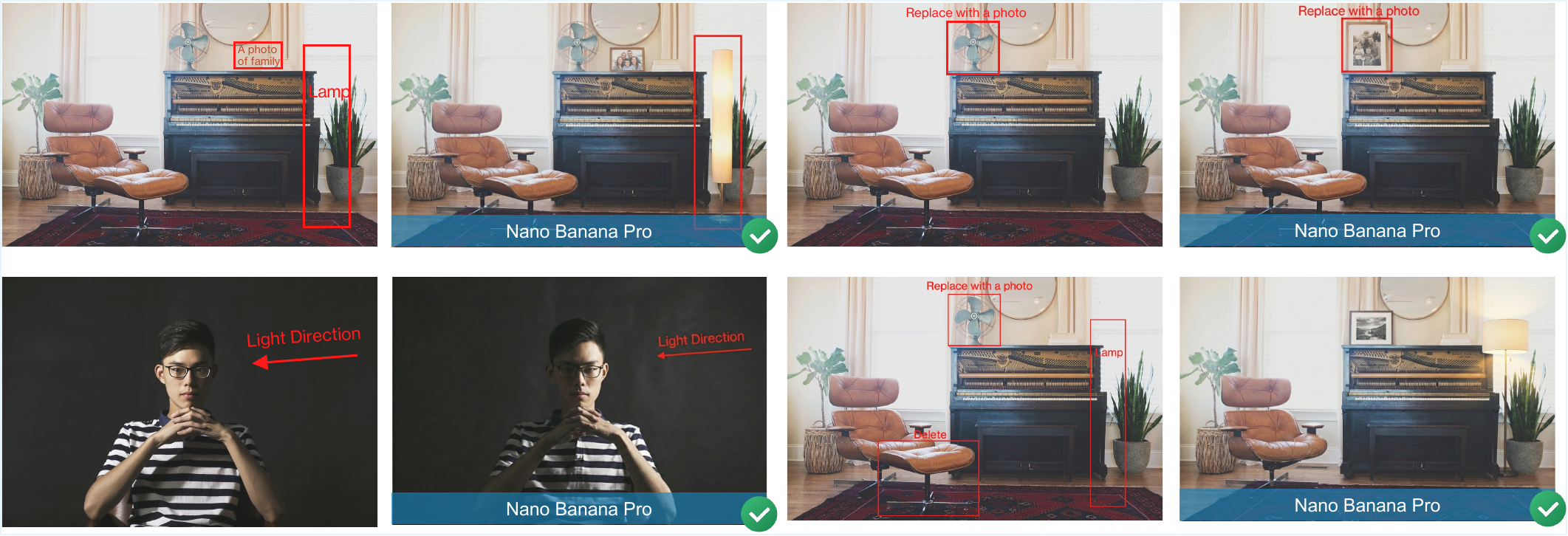}
\caption{\textbf{Qualitative examples with visually embedded instructions.}
All examples use the same minimal textual prompt, ``Edit this image following the instructions annotated on this picture.''
Task specifications are conveyed through text and symbols embedded directly in the input image.
Nano Banana Pro correctly executes single-task, multi-task, and causal editing operations based on these visually embedded instructions.}
    \label{fig:app_instruction}
\end{figure}

We further present qualitative case studies to examine whether models can follow instructions that are visually embedded within the input image.
In these experiments, the textual input is fixed to a minimal prompt,
\emph{``Edit this image following the instructions annotated on this picture.''}
All task specifications, including both textual descriptions and symbolic cues, are embedded directly into the image as visual annotations.

As shown in Figure~\ref{fig:app_instruction}, Nano Banana Pro demonstrates strong capability in interpreting and executing visually embedded instructions.
The model successfully handles both single-task and multi-task scenarios, including addition, removal, and replacement operations, as well as tasks involving causal reasoning such as light direction control.
These results indicate that the model is able to parse instruction content directly from the image, associate it with the corresponding operations, and apply the intended edits without relying on detailed textual prompts.

Notably, even in cases that require causal reasoning, such as modifying illumination according to an annotated light direction, the model produces results consistent with the embedded instructions.
This qualitative evidence suggests that frontier models can, to some extent, treat visually embedded instructions as primary guidance signals rather than auxiliary references.
Such behavior further motivates the need for systematic benchmarks like \abbr to characterize visual instruction-following capabilities beyond conventional text-centric prompting paradigms.

\begin{figure}
    \centering
    \includegraphics[width=0.7\linewidth]{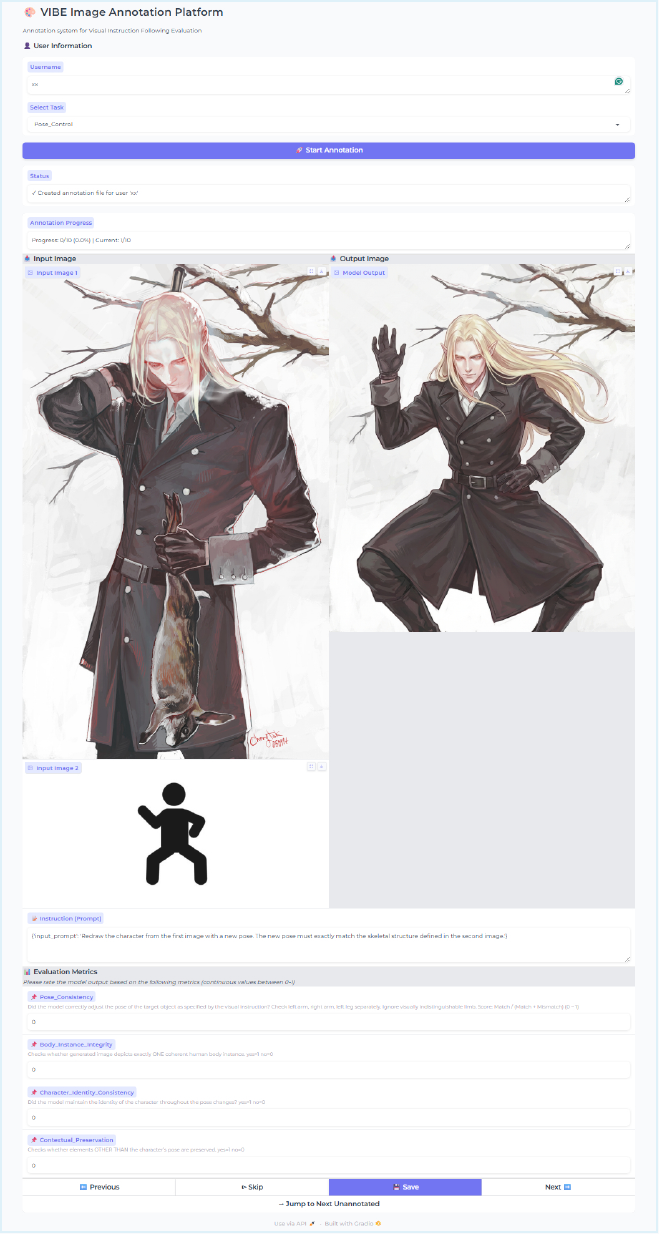}
\caption{\textbf{Screenshot of the developed data annotation system used in section~\ref{sec:eval_valid}.}}
    \label{fig:app_human}
\end{figure}
\section{Evaluation}
\label{app:eval}
\subsection{Evaluation Metrics}
As described in Section~\ref{sec:eval}, we design task-specific evaluation metrics tailored to the characteristics of each task.
Metrics for \textit{Addition}, \textit{Removal}, \textit{Replacement}, and \textit{Translation} are detailed in the main paper.
In this appendix, we provide the definitions of evaluation metrics for the remaining tasks.

\rrparagraph{Pose Control}
Evaluation for the \textit{Pose Control} task focuses on whether the target pose is correctly realized while preserving character integrity and contextual consistency.
We define four complementary criteria.
\begin{itemize}
    \item \textbf{Pose Consistency ($\mathcal{PC}$)} evaluates whether the target pose specified by the reference image is present in the generated output.
    This metric assesses pose correspondence regardless of whether the pose is realized through a physically valid or anatomically correct human body.
    The human body is decomposed into four coarse parts: left arm, right arm, left leg, and right leg.
    Each part is evaluated independently as a binary match against the reference pose.
    The final Pose Consistency score is computed as the average over the four parts, resulting in discrete values in $\{0, 0.25, 0.5, 0.75, 1\}$.

    \item \textbf{Body Instance Integrity ($\mathcal{BII}$)} explicitly evaluates whether the target pose is realized by a single coherent human body.
    It penalizes degenerate cases such as fragmented limbs, duplicated body parts, or pose realization through multiple inconsistent instances.
    $\mathcal{BII}$ is assigned a binary score.

    \item \textbf{Character Identity Consistency ($\mathcal{CIC}$)} measures whether the generated character remains identifiable as the same character as in the input image.
    This criterion evaluates the preservation of identity-related visual attributes and is scored binarily.

    \item \textbf{Contextual Preservation ($\mathcal{CP}$)} evaluates whether visual content outside the character’s pose remains unchanged.
    It penalizes unintended modifications to background elements or surrounding objects and is assigned a binary score.
\end{itemize}

The final score for the Pose Control task is computed as:
\begin{equation}
\text{Score} = (\mathcal{PC} \cdot \frac{\mathcal{BII} + \mathcal{CIC} + \mathcal{CP}}{ 3})^{\frac{1}{2}}.
\end{equation}

\rrparagraph{Reorientation}
The evaluation of the \textit{Reorientation} task focuses on whether the target object is correctly aligned to the specified orientation while preserving object identity and visual integrity.
We define three complementary metrics.
\begin{itemize}
    \item \textbf{Orientation Alignment ($\mathcal{OA}$)} evaluates whether the final orientation of the target object matches the target orientation specified by the reference indicator.
    The target orientation is defined along three independent axes: yaw, pitch, and roll.
    For each axis, a binary score is assigned based on whether the final object orientation is aligned with the target orientation along that axis.
    Axis-wise alignment is judged solely based on the final result, regardless of whether modification was required.
    The Orientation Alignment score is computed as the average of the three axis-wise scores, yielding discrete values in $\{0, \frac{1}{3}, \frac{2}{3}, 1\}$.

    \item \textbf{Identity Consistency ($\mathcal{IC}$)} evaluates whether the edited object in the generated image remains the same semantic entity as in the input image.
    This metric ignores changes directly induced by reorientation, such as pose, facing direction, or perspective.
    It penalizes object replacement, removal, duplication, or severe structural corruption.
    $\mathcal{IC}$ is assigned a binary score.

    \item \textbf{Visual Integrity ($\mathcal{VI}$)} evaluates whether the reorientation process introduces severe visual artifacts or layout corruption.
    This includes prominent visual pollution, large rendering artifacts, or structural image breakdown that significantly degrades readability.
    $\mathcal{VI}$ is assigned a binary score.
\end{itemize}

The final score for the Reorientation task is computed as the geometric mean of the three criteria:
\begin{equation}
\text{Score} = (\mathcal{OA} \cdot \frac{\mathcal{IC} + \mathcal{VI}}{2})^{\frac{1}{2}}.
\end{equation}

\rrparagraph{Draft Instantiation}
The evaluation protocol for the \textit{Draft Instantiation} task follows exactly the same metric design as that used for tasks in the Deictic Level.
Specifically, performance is assessed using Instruction Adherence ($\mathcal{IA}$), Contextual Preservation ($\mathcal{CP}$), and Visual Coherence ($\mathcal{VC}$), which is detailed in Section~\ref{sec:eval}.

\rrparagraph{Light Control}
The \textit{Light Control} task evaluates whether the generated image reflects the target lighting direction specified by an arrow while preserving non-lighting scene content.
We define two metrics: \textbf{Lighting Direction Consistency ($\mathcal{LDC}$)} and \textbf{Contextual Preservation ($\mathcal{CP}$)}.
\begin{itemize}
    \item \textbf{Lighting Direction Consistency ($\mathcal{LDC}$)} measures whether the dominant illumination direction on the subject in the generated image matches the target direction indicated by the arrow in the input image.
    $\mathcal{LDC}$ consists of two sub-metrics.
    \emph{Direction Matching Consistency ($\mathcal{DMC}$)} compares the arrow direction against the dominant lighting direction inferred from highlights and shadows on the subject (rather than from visible light sources).
    $\mathcal{DMC}$ is scored on a three-level scale: $1.0$ if the directions are nearly identical, $0.5$ if the lighting is modified toward the target direction but exhibits a noticeable angular deviation (within $\sim 90^\circ$), and $0.0$ if the lighting direction is largely different, unchanged, or ambiguous.
    \emph{Physical Lighting Consistency ($\mathcal{PLC}$)} evaluates whether the observed shading, shadowing, and highlight distribution are physically consistent with the target direction.
    $\mathcal{PLC}$ is only evaluated when $\mathcal{DMC}=1.0$; otherwise, $\mathcal{PLC}$ is set to $0$ by default.
    $\mathcal{PLC}$ is scored binarily: $1$ if the illumination pattern is physically consistent with the target direction, and $0$ otherwise.
    We compute $\mathcal{LDC}$ as the average of its two sub-metrics.
    
    \item \textbf{Contextual Preservation ($\mathcal{CP}$)} evaluates whether the generated image preserves all non-target content while applying the intended lighting-direction edit.
    $\mathcal{CP}$ is scored binarily and penalizes any unrelated content modifications, including object addition/removal, geometry or layout changes, and semantic alterations unrelated to lighting.
    $\mathcal{CP}$ is set to $1$ only if all observable differences between input and output are attributable to lighting-related effects; otherwise, $\mathcal{CP}$ is set to $0$.
\end{itemize}

The final score for the Light Control task is computed as the geometric mean of the two metrics:
\begin{equation}
\text{Score} = (\mathcal{LDC} \cdot \mathcal{CP})^{\frac{1}{2}}.
\end{equation}

\rrparagraph{Flow Simulation}
The \textit{Flow Simulation} task evaluates whether the generated image correctly reflects the target wind direction while preserving the identity and placement of wind-affected subjects.
We define two complementary metrics:

\begin{itemize}
    \item \textbf{Wind Direction Consistency ($\mathcal{WDC}$)} measures whether the dominant wind flow in the generated image aligns with the target direction specified by the arrow in the input image.
    Wind direction is inferred exclusively from visible, directionally consistent responses of wind-sensitive elements, such as hair, clothing, vegetation, smoke, particles, or flame shape.
    Abstract airflow cues without corresponding effects on scene elements are not considered valid evidence of wind.
    WDC is scored on a three-level scale: $1.0$ if at least one wind-sensitive element responds clearly and its motion closely matches the target direction; $0.5$ if wind effects are present and generally follow the target direction with a noticeable angular deviation (within $\sim 30^\circ$); and $0.0$ otherwise, including cases with no wind response, ambiguous motion, or inconsistent direction.

    \item \textbf{Contextual Preservation ($\mathcal{CP}$)} evaluates whether wind-affected subjects preserve their semantic identity and overall placement after editing, while allowing changes directly attributable to wind.
    $\mathcal{CP}$ consists of two sub-metrics.
    \emph{Wind-Identity Preservation ($\mathcal{WIP}$)} checks whether all wind-affected subjects remain the same semantic entities as in the input image, ignoring deformation or scattering caused by wind.
    \emph{Wind-Pose/Placement Preservation ($\mathcal{WPP}$)} evaluates whether the subjects’ global position and pose remain consistent after excluding wind-induced effects such as hair fluttering, cloth bending, or particle dispersal.
    $\mathcal{WPP}$ is evaluated only if $\mathcal{WIP}=1$; otherwise, it is set to $0$ by default.
    Both sub-metrics are scored binarily, and $\mathcal{CP}$ is computed as their average.
\end{itemize}

The final score for the Flow Simulation task is computed as the geometric mean of $\mathcal{WDC}$ and $\mathcal{CP}$:
\begin{equation}
\text{Score} = (\mathcal{WDC} \cdot \mathcal{CP})^{\frac{1}{2}}.
\end{equation}

\rrparagraph{Billiards}
The \textit{Billiards} task evaluates a model’s ability to reason about multi-step physical interactions, including ball motion, wall collisions, and target prediction.
Given a golden-label image depicting the correct initial state, future trajectory, and final target, the generated output is evaluated using three independent metrics.

\begin{itemize}
    \item \textbf{Path Correctness ($\mathcal{PC}$)} evaluates whether the predicted trajectory follows the same causal structure as the golden label.
    Rather than enforcing exact geometric overlap, this metric focuses on the topology and direction of motion.
    Specifically, it checks whether the trajectory proceeds in the same initial direction and whether it interacts with the same table cushions in the same order.
    Trajectories exhibiting incorrect wall collisions, reversed directions, or hallucinated loops are penalized.
    $\mathcal{PC}$ is assigned a binary score.

    \item \textbf{Collision Correctness ($\mathcal{CC}$)} evaluates whether the final target ball is correctly identified.
    This metric checks whether the predicted target region corresponds to the same ball number as specified in the golden label, regardless of minor deviations in the predicted path.
    $\mathcal{CC}$ is assigned a binary score and focuses exclusively on target identity rather than trajectory quality.

    \item \textbf{Contextual Preservation ($\mathcal{CP}$)} verifies whether the static environment remains unchanged.
    This metric checks whether all billiard balls are present, whether ball numbers are preserved, whether the spatial layout of non-moving balls is consistent with the golden label, and whether the directional arrow on the cue ball is correctly retained.
    Minor differences in lighting or rendering are ignored.
    $\mathcal{CP}$ is assigned a binary score and is set to $0$ if any ball is missing, added, renumbered, or significantly displaced, or if the directional arrow is incorrect.

\end{itemize}

The final score for the Billiards task is computed as follows:
\begin{equation}
    \text{Score} = (\frac{\mathcal{PC}+\mathcal{CC}}{2} \cdot \mathcal{CP})^{\frac{1}{2}}
\end{equation}

\subsection{Evaluation Prompt}
We provide the evaluation prompts of each metric in Table~\ref{tab:app_prompt_ia} $\sim$ \ref{tab:app_prompt_reo_rcp}.

\begin{table*}[]
\centering
\caption{Evaluation Prompt for Instruction Adherence}
\begin{tcolorbox}[title = {\textbf{Instruction Adherence}}]

You are given THREE images and ONE text prompt.

The first image:

- This is the original image.

The second image:

- Visual instructions are drawn on this image based on the original image by the user.

The third image:

- This is the image generated by a model after editing the second image.
\\

TEXT PROMPT:

- \{prompt\}
\\

Your task is to evaluate whether the Output Image (The third image) correctly follows the instruction.
The instruction consists of BOTH:\\
(1) the visual instruction drawn on the second image, and\\
(2) the textual description in the Text Prompt.
\\
You must independently evaluate the following THREE metrics and assign a binary score (1 or 0) to each:\\
\\
1. Visual Instruction Localization Correctness  
Did the main edit occur on the object or region explicitly indicated by the visual instruction on the Input Image (The second image)?\\
2. Visual Operator Type Compliance  
Was the type of edit consistent with the operation implied by the visual instruction?\\
3. Textual Action Semantic Compliance  
Did the model execute the core action specified in the Text Prompt?\\
\\
Scoring rules:\\
- Score 1 if the requirement is clearly satisfied.\\
- Score 0 if the requirement is not satisfied or is ambiguous.\\
- If unsure, assign 0.\\
- Partial compliance must be scored as 0.\\
\\
You may reason freely to reach your decision. \\
Then, for EACH metric, provide:\\
- "score": an integer value of 0 or 1.\\
- "reason": ONE short factual sentence describing an observable outcome.\\
This summary must strictly follow the output format specified below:\\
\\
\{\\
  ``Visual\_Instruction\_Localization\_Correctness": \{\\
    ``reason": "brief factual summary",\\
    ``score": 1/0\\
  \},\\
  ``Visual\_Operator\_Type\_Compliance": \{\\
    ``reason": "brief factual summary",\\
    ``score": 1/0\\
  \},\\
  ``Textual\_Action\_Semantic\_Compliance": \{\\
    ``reason": "brief factual summary",\\
    ``score": 1/0\\
  \}\\
\}

\end{tcolorbox}
\label{tab:app_prompt_ia}
\end{table*}

\begin{table*}[]
\centering
\caption{Evaluation Prompt for Contextual Preservation 1/2}
\begin{tcolorbox}[title = {\textbf{Contextual Preservation 1/2}}]
You are given TWO images and ONE text prompt.\\

IMAGE 1 (Input Image):\\
- Original image with user-drawn visual instructions.\\

IMAGE 2 (Output Image):\\
- Image generated by the model after editing IMAGE 1.\\
- The output may be cropped or reframed.\\

TEXT PROMPT:\\
- \{prompt\}\\

Your task is to evaluate Contextual Preservation.\\

Definition:\\
Contextual Preservation checks whether the model changed anything it was NOT supposed to change.\\
It does NOT judge whether the edit was correct, precise, or well aligned.\\
Errors in edit location, extent, or alignment belong to Instruction Adherence, not Contextual Preservation.\\

Evaluation rules (follow strictly):\\

1) Cropping rule  \\
- If the output is cropped, only compare the overlapping visible region.\\
- Ignore content missing only due to cropping.\\

2) Difference listing (what counts as a difference)\\
- List ONLY meaningful differences at the level of objects or semantic entities.\\
- Do NOT list differences caused by:\\
  • minor blur or softness,\\
  • small texture or color shifts,\\
  • pixel-level noise,\\
  • slight position or alignment offsets.\\
- A difference should be listed ONLY if it:\\
  • adds or removes a complete object,\\
  • changes the identity of an object,\\
  • or damages the structural integrity of a non-target object.\\
  
3) Target rule  \\
- Identify the intended edit target based ONLY on:\\
  (a) the visual instruction marks, and\\
  (b) the text prompt.\\

4) Classification rule\\
- IN\_TARGET:
 - IN\_TARGET:\\
  • any change within the intended target,\\
  • OR any imperfect attempt to edit the target\\
    (including misplacement, offset, scale error, or incomplete coverage).\\

\end{tcolorbox}
\label{tab:app_prompt_cp1}
\end{table*}

\begin{table*}[]
\centering
\caption{Evaluation Prompt for Contextual Preservation 2/2}
\begin{tcolorbox}[title = {\textbf{Contextual Preservation 2/2}}]
- OUT\_OF\_TARGET:\\
  • any change to unrelated objects or regions,\\
  • any addition or removal of unrelated semantic entities,\\
  • any structural damage to non-target objects.\\

5) Scoring  \\
- Score = 1 if NO OUT\_OF\_TARGET differences exist.\\
- Score = 0 if ANY OUT\_OF\_TARG\\

Output format:\\
First provide a brief analysis with these sections:\\
- \#\# Differences\\
- \#\# Target\\
- \#\# Classification\\
- \#\#  Decision\\

Then output the final JSON as the last part of your response:\\

\{
  ``Contextual\_Preservation": \{
    ``reason": ``string",
    ``score": 0
  \}
\}

5) Scoring  \\
- Score = 1 if NO OUT\_OF\_TARGET differences exist.\\
- Score = 0 if ANY OUT\_OF\_TARG\\

\end{tcolorbox}
\label{tab:app_prompt_cp2}
\end{table*}

\begin{table*}[]
\centering
\caption{Evaluation Prompt for Visual Coherence 1/2}
\begin{tcolorbox}[title = {\textbf{Visual Coherence 1/2}}]
You are given THREE images and ONE text prompt.\\

The first image:\\
- This is the original image.\\

The second image:\\
- Visual instructions are drawn on this image based on the original image by the user.\\

The third image:\\
- This is the image generated by a model after editing the second image.\\

TEXT PROMPT:\\
- \{prompt\}\\

Your task is to evaluate the Visual Coherence of the Output Image.\\

Visual Coherence evaluates whether the edited result is visually unified and generatively sound.\\
This metric is STYLE-AGNOSTIC: you must NOT judge realism, beauty, or artistic preference.\\
Instead, you must assess whether the edited image remains consistent with the source image and whether the output avoids clear generative artifacts.\\

You must independently evaluate the following THREE metrics and assign a binary score (1 or 0) to each:\\

1. Style Consistency\\
Did the edited region adopt the same artistic or rendering domain as the Input Image\\
(e.g., line-art, watercolor, oil painting, 3D render, photographic style, pixel art, animation)?\\

- Score 1 if the edited region clearly belongs to the same visual domain as the source image.\\
- Score 0 if the edited region introduces a different artistic or rendering domain.\\
- Do NOT judge whether the style looks good or realistic---only whether it matches the source domain.\\

2. Visual Seamlessness\\
Is the edited region visually continuous with its surrounding area, without obvious signs of compositing?\\

Focus on whether there are clear visual discontinuities such as:\\
- unnatural seams or hard boundaries,\\
- abrupt changes in texture, color, or resolution,\\
- visible cut-and-paste artifacts.\\

- Score 1 if the edited region integrates seamlessly with its surroundings.\\
- Score 0 if there are clear and noticeable discontinuities.\\
- Do NOT penalize stylistic roughness if it is consistent with the source image.\\

3. Artifact-Free Generation\\
Does the Output Image avoid obvious, domain-independent generative artifacts?\\

\end{tcolorbox}
\label{tab:app_prompt_vc1}
\end{table*}

\begin{table*}[]
\centering
\caption{Evaluation Prompt for Visual Coherence 2/2}
\begin{tcolorbox}[title = {\textbf{Visual Coherence 2/2}}]
Consider artifacts such as:\\
- unintended blurring or pixelation,\\
- geometric distortion or deformation,\\
- broken structures, duplicated patterns, or rendering collapse,\\
- residual visual instruction marks such as arrows, boxes, strokes, or masks that should not appear in the final image.\\

- Score 1 if no clear generative artifacts are present.\\
- Score 0 if obvious, non-stylistic artifacts are visible.\\
- Style-specific noise or abstraction is allowed if consistent with the source domain.\\

General rules:\\
- Evaluate ONLY visual coherence, not instruction correctness or contextual preservation.\\
- Minor pixel-level differences are acceptable.\\
- If you are unsure about a case, assign score 0.\\
- Partial compliance must be scored as 0.\\

You may reason freely to reach your decision.\\
Then, for EACH metric, you provide a summary:\\
- ``score'': an integer value of 0 or 1.\\
- ``reason'': ONE short factual sentence describing an observable outcome.\\

This summary must strictly follow the output format specified below:\\

\{
\quad ``Style\_Consistency'': \{\\
\quad\quad ``reason'': ``brief factual summary'',\\
\quad\quad ``score'': 1/0\\
\quad \},\\
\quad ``Visual\_Seamlessness'': \{\\
\quad\quad ``reason'': ``brief factual summary'',\\
\quad\quad ``score'': 1/0\\
\quad \},\\
\quad ``Artifact-Free\_Generation'': \{\\
\quad\quad ``reason'': ``brief factual summary'',\\
\quad\quad ``score'': 1/0\\
\quad \}\\
\}
\end{tcolorbox}
\label{tab:app_prompt_vc2}
\end{table*}

\begin{table*}[]
\centering
\caption{Evaluation Prompt for Billiards 1/2}
\begin{tcolorbox}[title = {Billiards 1/2}]
You are an expert computer vision evaluator for physics simulation models.\\

You are provided with two images:\\
1. IMAGE 1 (Golden Label / Ground Truth): Depicts the correct initial state, the correct future trajectory (green dashed line), and the correct final target (red bounding box).\\
2. IMAGE 2 (Model Generation): Depicts a predicted scene, trajectory, and target.\\

YOUR TASK:\\
Evaluate the Generated Image against the Golden Label based on three independent metrics. You must analyze the images step-by-step and output a final JSON score.\\

1. Context Preservation (CP)\\
Goal: Verify that the static environment remains unchanged.\\
Check:
- Are all billiard balls (balls with numbers and the white ball) present?\\
- Are the ball numbers consistent?\\
- Is the spatial layout (positions of non-moving balls) consistent with the ground truth?\\
- Is the black arrow on the white ball preserved?\\
Ignore: Slight changes in lighting, shading, or minor pixel-level rendering differences.\\
Scoring:
- 0: If any ball is missing, added, re-numbered, or significantly displaced; or if the arrow is missing or incorrect.\\
- 1: If the static scene identity and layout are preserved.\\

2. Path Correctness (PC)\\
Goal: Verify the topology and direction of the green dashed trajectory.\\
Check:
- Does the predicted path move in the same cardinal direction?\\
- Does the trajectory bounce off the same specific walls or cushions in the same order?\\
  (e.g., if Truth hits Top-Wall then Left-Wall, Prediction must do the same).\\
- Is the path free of hallucinations (e.g., random loops or squiggly lines)?\\
Scoring:
- 0: If the trajectory hits different walls, moves in a different initial direction, or has a completely different shape.\\
- 1: If the trajectory follows the same sequence of wall impacts and general geometry.\\

3. Collision Correctness (CC)\\
Goal: Verify the final target identity.\\
Check:
- Does the red bounding box surround the same specific ball number as in the Golden Label?\\
Note: This metric is strictly about the identity of the target, regardless of whether the path (PC) looks perfect.\\
Scoring:
- 0: If the red box highlights a different ball or an empty space.\\
- 1: If the red box highlights the correct ball number.\\

Output Instructions\\
Step 1: Analysis \& Decisions\\
Provide a text analysis under the headers \#\# Analysis and \#\# Decisions.\\
For each metric, explicitly state the observable differences or similarities.\\
Example for Path: ``The Golden path bounces off the top cushion. The Generated path bounces off the bottom cushion. These are different.''\\

Step 2: Final JSON\\
Output the final score in the following JSON format. This must be the last part of your response.\\

\end{tcolorbox}
\label{tab:app_prompt_bil1}
\end{table*}

\begin{table*}[]
\centering
\caption{Evaluation Prompt for Billiards 2/2}
\begin{tcolorbox}[title = {Billiards 2/2}]

\{
\quad ``Context\_Preservation'': \{\\
\quad\quad ``reason'': ``One sentence summary focusing on ball count, numbers, and layout.'',\\
\quad\quad ``score'': 0 or 1\\
\quad \},\\
\quad ``Path\_Correctness'': \{\\
\quad\quad ``reason'': ``One sentence summary focusing on trajectory direction and wall bounce sequence.'',\\
\quad\quad ``score'': 0 or 1\\
\quad \},\\
\quad ``Collision\_Correctness'': \{\\
\quad\quad ``reason'': ``One sentence summary confirming if the correct ball number was targeted.'',\\
\quad\quad ``score'': 0 or 1\\
\quad \}\\
\}
\end{tcolorbox}
\label{tab:app_prompt_bil2}
\end{table*}

\begin{table*}[]
\centering
\caption{Evaluation Prompt for Wind Contextual Preservation 1/2}
\begin{tcolorbox}[title = {Wind Contextual Preservation 1/2}]
You are given TWO images.\\

IMAGE 1 (Input Image):\\
- The original image before wind editing.\\
- This image may contain visual instruction markings\\
  (e.g., arrows) indicating wind direction.\\

IMAGE 2 (Generated Image):\\
- The model-generated image after wind editing.\\

Your task is to evaluate Wind Contextual Preservation (W-CP), which consists of TWO sub-metrics:\\
1) Wind-Identity Preservation (W-IP)\\
2) Wind-Pose/Placement Preservation (W-PP)\\

IMPORTANT GLOBAL RULES:\\
- Visual instruction markings (e.g., arrows) are NOT scene content. Differences related to arrows MUST be ignored, unless they severely obstruct the subject or degrade visual quality.\\
- Wind effects are assumed to be intentional and allowed.\\
- Do NOT penalize changes directly caused by wind.\\

------------------------------------------------------------\\
SUB-METRIC 1: Wind-Identity Preservation (W-IP)\\

W-IP checks whether the wind-affected subject(s) remain the SAME semantic entity after editing.\\

Focus on identity ONLY:\\
- Person identity (same person, same character)\\
- Object identity (same candle, same dandelion, same clothing item)\\

Explicitly IGNORE:\\
- Deformation, bending, or scattering caused by wind (e.g., flying hair, dispersed seeds, flickering flame).\\

Scoring:\\
- Score = 1\\
  If the identity of all wind-affected subjects is preserved.\\

- Score = 0\\
  If any subject is removed, replaced, duplicated, or transformed into a different semantic entity.\\

------------------------------------------------------------\\
SUB-METRIC 2: Wind-Pose / Placement Preservation (W-PP)\\

IMPORTANT:\\
- Evaluate W-PP ONLY IF W-IP = 1.\\
- If W-IP = 0, set W-PP = 0 by default.\\

W-PP checks whether the overall position and pose of the subject(s) remain consistent with IMAGE 1, AFTER ignoring wind-induced effects.\\

\end{tcolorbox}
\label{tab:app_prompt_fs_wcp1}
\end{table*}

\begin{table*}[]
\centering
\caption{Evaluation Prompt for Wind Contextual Preservation 2/2}
\begin{tcolorbox}[title = {Wind Contextual Preservation 2/2}]

Ignore the following wind-induced changes:\\
- Hair blowing direction or shape.\\
- Cloth fluttering or bending.\\
- Dispersal or motion of lightweight elements (e.g., dandelion seeds, smoke, flame shape).\\

Check ONLY:\\
- Whether the subject’s main body has shifted position.\\
- Whether the subject’s global orientation or stance has changed.\\

Scoring:\\
- Score = 1\\
  If the subject’s overall position and pose are preserved.\\

- Score = 0\\
  If the subject has been significantly moved, rotated, or reposed beyond what wind could plausibly cause.\\

------------------------------------------------------------\\
Output Requirements:\\

1) First provide your reasoning under:\\
   - \#\# Identity Preservation Analysis\\
   - \#\# Pose / Placement Analysis\\

2) Then output the final JSON under:\\
   - \#\# JSON\\

3) The JSON must be the LAST part of your response.\\

------------------------------------------------------------\\
FINAL JSON FORMAT (EXACT):\\

\{
\quad ``Wind-Identity\_Preservation'': \{\\
\quad\quad ``reason'': ``string'',\\
\quad\quad ``score'': 0\\
\quad \},\\
\quad ``Wind-Other\_Preservation'': \{\\
\quad\quad ``reason'': ``string'',\\
\quad\quad ``score'': 0\\
\quad \}\\
\}
\end{tcolorbox}
\label{tab:app_prompt_fs_wcp2}
\end{table*}

\begin{table*}[]
\centering
\caption{Evaluation Prompt for Wind Direction Consistency 1/2}
\begin{tcolorbox}[title = {Wind Direction Consistency 1/2}]
You are given TWO images.\\

IMAGE 1 (Input Image):\\
- The original image.\\
- A visual arrow is drawn on this image indicating the desired wind direction.\\
- The arrow defines the TARGET wind flow direction.\\

IMAGE 2 (Generated Image):\\
- The model-generated image after wind editing.\\

Your task is to evaluate Wind Direction Consistency (WDC).\\

IMPORTANT SCOPE:\\
- Evaluate ONLY the wind direction.\\
- Do NOT evaluate physical realism, wind strength, or scene preservation.\\
- Ignore all non-wind-related changes.\\

------------------------------------------------------------\\
CRITICAL CAUSAL RULE (MUST FOLLOW):\\

Wind direction is considered valid ONLY IF it produces\\
visible, directionally consistent effects on at least one\\
wind-sensitive element in the scene.\\

Wind-sensitive elements include (but are not limited to):\\
- Hair\\
- Clothing or fabric\\
- Vegetation (e.g., trees, grass, dandelions)\\
- Smoke, mist, particles\\
- Flame shape (e.g., candles)\\

If wind is shown only as abstract airflow (e.g., lines or particles)\\
WITHOUT affecting any wind-sensitive element,\\
the wind direction MUST be scored as 0.\\

------------------------------------------------------------\\
PART 1: Direction description (no comparison)\\

(1) Arrow direction in IMAGE 1:\\
Describe the arrow direction as a continuous spatial direction.\\
Do NOT reduce it to simple categories (left/right/up/down).\\
Use precise descriptions (e.g., ``from upper-left toward lower-right'').\\

(2) Wind direction in IMAGE 2:\\
Infer the dominant wind direction based ONLY on visible wind effects,\\
such as hair, clothing, vegetation, smoke, or particles.\\
Describe the wind direction using the same level of precision as above.\\

\end{tcolorbox}
\label{tab:app_prompt_fs_wdc1}
\end{table*}

\begin{table*}[]
\centering
\caption{Evaluation Prompt for Wind Direction Consistency 2/2}
\begin{tcolorbox}[title = {Wind Direction Consistency 2/2}]
------------------------------------------------------------\\
PART 2: Direction comparison and scoring\\

Compare the arrow direction and the inferred wind direction.\\

Scoring:\\
- Score = 1.0\\
  ONLY if:\\
  (a) At least one wind-sensitive element clearly responds to wind, AND\\
  (b) The response direction closely matches the arrow direction.\\

- Score = 0.5\\
  ONLY if:\\
  (a) Wind-sensitive elements respond to wind, AND\\
  (b) The response direction generally follows the arrow, but with a noticeable angular deviation (within $\sim$30 degrees).\\

- Score = 0.0\\
  In ALL other cases, including:\\
  - No wind-sensitive elements respond to wind.\\
  - Wind is shown only via abstract airflow cues.\\
  - Wind direction is unchanged, opposite, or ambiguous.\\
  - Wind-sensitive elements exist but do not move consistently\\
    with the arrow direction.\\

------------------------------------------------------------\\
Output Requirements:\\

1) First provide your reasoning under:\\
   - \#\# Causal Wind Evidence\\
   - \#\# Direction Description\\
   - \#\# Direction Comparison\\

2) Then output the final JSON under:\\
   - \#\# JSON\\

3) The JSON must be the LAST part of your response.\\

------------------------------------------------------------\\
FINAL JSON FORMAT (EXACT):\\

\{
\quad ``Wind\_Direction\_Consistency'': \{\\
\quad\quad ``reason'': ``string (one-sentence summary)'',\\
\quad\quad ``score'': 0.0\\
\quad \}\\
\}
\end{tcolorbox}
\label{tab:app_prompt_fs_wdc2}
\end{table*}

\begin{table*}[]
\centering
\caption{Evaluation Prompt for Contextual Preservation in Light Control 1/2}
\begin{tcolorbox}[title = {Contextual Preservation in Light Control 1/2}]
You are given TWO images.\\

IMAGE 1 (Input Image):\\
- The original image before editing.\\

IMAGE 2 (Generated Image):\\
- The model-generated image after lighting editing.\\

Your task is to evaluate Contextual Preservation (CP).\\

CP checks whether IMAGE 2 preserves all non-target content from IMAGE 1 while applying the intended lighting-direction edit.\\

IMPORTANT SCOPE:\\
- This evaluation is NOT about lighting direction correctness.\\
- Assume lighting direction editing is the intended and allowed operation.\\
- CP ONLY checks whether unrelated scene content has been modified.\\

--- Visual Instruction Rule:\\
- Visual instruction markings in IMAGE 1 (e.g., arrows or guides) are NOT part of the scene content.\\
- Differences related to the presence, absence, or appearance of these instruction markings in IMAGE 2 MUST be IGNORED, UNLESS they severely obstruct the subject or significantly degrade the visual quality of the image.\\

------------------------------------------------------------\\
PART 1: Allowed vs non-allowed changes\\

Allowed changes (do NOT penalize):\\
- Changes in illumination intensity, shading, or highlights.\\
- Changes in shadow shape or placement caused by lighting.\\
- Minor color shifts attributable to lighting.\\
- Appearance of visible light sources or light-related effects introduced solely to express lighting, such as:\\
  • bright light spots,\\
  • visible lamps,\\
  • sun glare,\\
  • lens flare,\\
  • light beams or glow effects, as long as they do not alter scene geometry or object structure.\\

Non-allowed changes (must be penalized):\\
- Addition or removal of non-lighting objects.\\
- Changes in object geometry or structure.\\
- Changes in scene layout or spatial arrangement.\\
- Replacement of materials, textures, or clothing.\\
- Introduction of new semantic scene elements unrelated to lighting (e.g., new furniture, new buildings).\\
- Differences involving visual instruction markings (e.g., arrows) that do not affect scene content.

------------------------------------------------------------\\
PART 2: Difference-based comparison\\

Compare IMAGE 1 and IMAGE 2 carefully.\\
\end{tcolorbox}
\label{tab:app_prompt_lc_cp1}
\end{table*}

\begin{table*}[]
\centering
\caption{Evaluation Prompt for Contextual Preservation in Light Control 2/2}
\begin{tcolorbox}[title = {Contextual Preservation in Light Control 2/2}]
Procedure:\\
1) List all observable differences between IMAGE 1 and IMAGE 2.\\
2) For each difference, determine whether it is:\\
   - a lighting-related manifestation (allowed), OR\\
   - an unrelated content modification (not allowed).\\

------------------------------------------------------------\\
PART 3: Contextual Preservation decision\\

Scoring (binary):\\
- Score = 1\\
  ONLY if ALL observed differences fall under allowed lighting-related changes.\\

- Score = 0\\
  If ANY difference corresponds to an unrelated content modification.\\

If unsure whether a difference is lighting-related or not, treat it as NOT preserved (score 0).\\

------------------------------------------------------------\\
Output Requirements:\\

1) First provide your reasoning under:\\
   - \#\# Difference Analysis\\
   - \#\# CP Decision\\

2) Then output the final JSON under:\\
   - \#\# JSON\\

3) The JSON must be the LAST part of your response.\\

------------------------------------------------------------\\
FINAL JSON FORMAT (EXACT):\\

\{
\quad ``Contextual\_Preservation'': \{\\
\quad\quad ``reason'': ``string (one-sentence summary)'',\\
\quad\quad ``score'': 0\\
\quad \}\\
\}
\end{tcolorbox}
\label{tab:app_prompt_lc_cp2}
\end{table*}

\begin{table*}[]
\centering
\caption{Evaluation Prompt for Lighting Direction Consistency 1/2}
\begin{tcolorbox}[title = {Lighting Direction Consistency 1/2}]
You are given TWO images.\\

IMAGE 1 (Input Image):\\
- The original image.\\
- A visual arrow indicates the desired lighting direction.\\
- The arrow defines the TARGET light propagation direction (from the light source toward the subject).\\

IMAGE 2 (Generated Image):\\
- The model-generated image after lighting editing.\\

Your task is to evaluate Lighting Direction Consistency (LDC), which consists of TWO sub-metrics:\\
1) Direction Matching Consistency (DMC)\\
2) Physical Lighting Consistency (PLC)\\

------------------------------------------------------------\\
PART 1: Direction description (no comparison)\\

(1) Arrow direction in IMAGE 1:\\
Describe the arrow direction as a continuous spatial direction.\\
Use precise language (e.g., ``from upper-right toward lower-left'', ``slightly downward from right to left'').\\
Do NOT reduce the direction to simple categories like left/right/up/down.\\

(2) Lighting direction on the subject in IMAGE 2:\\
Describe the dominant lighting direction ON THE SUBJECT, inferred ONLY from surface illumination (highlights and shadows), not from light source position or visible beams.\\
Use the same level of directional precision as above.\\

------------------------------------------------------------\\
PART 2: Direction Matching Consistency (DMC)\\

Compare the two described directions.\\

Scoring:\\
- Score = 1.0\\
  If the directions are nearly identical with no clear angular deviation.\\

- Score = 0.5\\
  If the lighting direction is clearly modified to follow the arrow, but exhibits a noticeable angular deviation (within $\sim$90 degrees).\\

- Score = 0.0\\
  If the lighting direction is largely different, unchanged, or ambiguous.\\

------------------------------------------------------------\\
PART 3: Physical Lighting Consistency (PLC)\\

IMPORTANT:\\
- Only evaluate PLC IF DMC = 1.0.\\
- If DMC < 1.0, set PLC = 0 by default.\\

\end{tcolorbox}
\label{tab:app_prompt_lc_ldc1}
\end{table*}

\begin{table*}[]
\centering
\caption{Evaluation Prompt for Lighting Direction Consistency 2/2}
\begin{tcolorbox}[title = {Lighting Direction Consistency 2/2}]

Assume the arrow direction is the TRUE light source direction.\\
Check whether the lighting effects on the subject in IMAGE 2 are physically consistent with that direction.\\

Consider:\\
- Which side of the subject should be illuminated.\\
- Which side should be in shadow.\\
- Whether highlights and shading agree with the arrow direction.\\

Scoring:\\
- Score = 1\\
  If the observed illumination pattern on the subject is physically consistent with the arrow direction.\\

- Score = 0\\
  If the illumination pattern contradicts the arrow direction.\\

------------------------------------------------------------\\
Output Requirements:\\

1) First provide your reasoning under:\\
   - \#\# Direction Description\\
   - \#\# Direction Comparison\\
   - \#\# Physical Consistency\\

2) Then output the final JSON under:\\
   - \#\# JSON\\

3) The JSON must be the LAST part of your response.\\

------------------------------------------------------------\\
FINAL JSON FORMAT (EXACT):\\

\{
\quad ``Direction\_Matching\_Consistency'': \{\\
\quad\quad ``reason'': ``string'',\\
\quad\quad ``score'': 0.0\\
\quad \},\\
\quad ``Physical\_Lighting\_Consistency'': \{\\
\quad\quad ``reason'': ``string'',\\
\quad\quad ``score'': 0\\
\quad \}\\
\}
\end{tcolorbox}
\label{tab:app_prompt_lc_ldc 2/2}
\end{table*}

\begin{table*}[]
\centering
\caption{Evaluation Prompt for BII, CIC, and Context Preservation 1/2}
\begin{tcolorbox}[title = {BII, CIC, and Context Preservation 1/2}]
You are given TWO images.\\

IMAGE 1 (Original Image):\\
- The original character image.\\

IMAGE 2 (Generated Image):\\
- The model-generated image.\\

Your task is to evaluate THREE independent metrics:\\
1) Body Instance Integrity (BII)\\
2) Character Identity Consistency (CIC)\\
3) Context Preservation (CP)\\

IMPORTANT:\\
- Do NOT evaluate pose correctness.\\
- Do NOT compare limb positions or body posture.\\
- Each metric must be judged independently.\\

------------------------------------------------------------\\
Metric definitions:\\

(1) Body Instance Integrity (BII)\\
Checks whether IMAGE 2 depicts exactly ONE coherent human body instance.\\

Score 0 if ANY of the following occur:\\
- Extra or duplicated limbs (e.g., more than two arms or legs).\\
- Multiple bodies, torsos, or heads.\\
- Visible skeletons, stick figures, or pose-reference structures overlaid on the body.\\
- Limbs are completely missing or not visible at all\\
  (unless they are plausibly occluded by clothing such as long skirts).\\

Otherwise, score 1.\\

------------------------------------------------------------\\
(2) Character Identity Consistency (CIC)\\
Checks whether the character in IMAGE 2 is the SAME character as in IMAGE 1.\\

Focus on:\\
- Face and facial features (if visible).\\
- Hair style and hair color.\\
- Distinctive identity cues (e.g., glasses, accessories).\\

Ignore:\\
- Pose changes.\\
- Background changes.\\
- Minor rendering differences.\\

Score 1 only if identity clearly matches.\\
If identity is different or cannot be confidently verified, score 0.\\

\end{tcolorbox}
\label{tab:app_prompt_pc_bcc1}
\end{table*}

\begin{table*}[]
\centering
\caption{Evaluation Prompt for BII, CIC, and Context Preservation 2/2}
\begin{tcolorbox}[title = {BII, CIC, and Context Preservation 2/2}]

------------------------------------------------------------\\
(3) Context Preservation (CP)\\
Checks whether elements OTHER THAN the character’s pose are preserved.\\

Focus on:\\
- Background scene.\\
- Clothing (style and type).\\
- Overall visual domain (e.g., realistic, anime).\\

Ignore:\\
- Natural cloth deformation caused by pose change.\\
- Minor occlusions or shading changes.\\

Score 0 if background, clothing, or scene elements are added, removed, or replaced.\\
Otherwise, score 1.\\

------------------------------------------------------------\\
Output requirements:\\

1) First provide your reasoning under these headers:\\
   - \#\# Analysis\\
   - \#\# Decisions\\

2) Then output the final JSON under:\\
   - \#\# JSON\\

3) Each metric must include:\\
   - ``reason": one short factual sentence\\
   - ``score": 0 or 1\\

4) The JSON must be the LAST part of your response.\\

------------------------------------------------------------\\
FINAL JSON FORMAT (EXACT):\\

\{\\
\ \ ``Body Instance Integrity": \{\\
\ \ \ \ ``reason": ``string (one-sentence summary)",\\
\ \ \ \ ``score": 0/1\\
\ \ \},\\
\ \ ``Character Identity Consistency": \{\\
\ \ \ \ ``reason": ``string (one-sentence summary)",\\
\ \ \ \ ``score": 0/1\\
\ \ \},\\
\ \ ``Context Preservation": \{\\
\ \ \ \ ``reason": ``string (one-sentence summary)",\\
\ \ \ \ ``score": 0/1\\
\ \ \}\\
\}\\

\end{tcolorbox}
\label{tab:app_prompt_pc_bcc2}
\end{table*}

\begin{table*}[]
\centering
\caption{Evaluation Prompt for Pose Consistency 1/2}
\begin{tcolorbox}[title = {Pose Consistency 1/2}]
You are given TWO images.\\

IMAGE 1 (Pose Reference):\\
- A pose schematic image (e.g., stick figure, skeleton, or line drawing).\\
- This image defines the TARGET POSE only.\\

IMAGE 2 (Generated Image):\\
- A model-generated image containing a character.\\

Your task is to evaluate POSE CONSISTENCY by comparing ONLY the limb poses.\\

IMPORTANT SCOPE:\\
- Ignore character identity, appearance, clothing, realism, or visual style.\\
- The character in IMAGE 2 may be a full human, skeleton, or line drawing.\\
- ONLY compare limb pose.\\
- Do NOT evaluate head or torso pose.\\

------------------------------------------------------------\\
OUTPUT STRUCTURE (must follow exactly):\\

PART 1: Limb pose description (no comparison)\\

Describe the limb poses in IMAGE 1 (pose reference):\\
- Left Arm:\\
- Right Arm:\\
- Left Leg:\\
- Right Leg:\\

Describe the limb poses in IMAGE 2 (generated image):\\
- Left Arm:\\
- Right Arm:\\
- Left Leg:\\
- Right Leg:\\

------------------------------------------------------------\\
PART 2: Limb-by-limb comparison\\

For each limb, compare IMAGE 2 against IMAGE 1 and assign ONE label:\\

- MATCH:\\
  Limb pose matches the reference in orientation, bending, and functional role.\\

- MISMATCH:\\
  Limb is visible but clearly differs from the reference (wrong bending, wrong pose, or incorrect side).\\

- N/A:\\
  Limb cannot be reliably evaluated because it is fully occluded, cropped out, or visually indistinguishable.\\

\end{tcolorbox}
\label{tab:app_prompt_pc_pc1}
\end{table*}

\begin{table*}[]
\centering
\caption{Evaluation Prompt for Pose Consistency 2/2}
\begin{tcolorbox}[title = {Pose Consistency 2/2}]

Rules:\\
- Minor spatial offsets and scale differences are allowed.\\
- If unsure, choose MISMATCH.\\
- Do NOT infer pose for limbs marked as N/A.\\

------------------------------------------------------------\\
PART 3: Final JSON output\\

Output the comparison results as JSON.\\
The JSON must be the LAST part of your response.\\

------------------------------------------------------------\\
FINAL JSON FORMAT (EXACT):\\

\{\\
\ \ ``Pose\_Consistency": \{\\
\ \ \ \ ``Left\_Arm": ``MATCH | MISMATCH | N/A",\\
\ \ \ \ ``Right\_Arm": ``MATCH | MISMATCH | N/A",\\
\ \ \ \ ``Left\_Leg": ``MATCH | MISMATCH | N/A",\\
\ \ \ \ ``Right\_Leg": ``MATCH | MISMATCH | N/A"\\
\ \ \}\\
\}\\

\end{tcolorbox}
\label{tab:app_prompt_pc_pc2}
\end{table*}

\begin{table*}[]
\centering
\caption{Evaluation Prompt for Orientation Alignment 1/3}
\begin{tcolorbox}[title = {Orientation Alignment 1/3}]
You are given TWO images.\\

\#\# IMAGE 1 (Input Image)\\
- The original image.\\
- A FOUR-SIDED PYRAMID orientation indicator is shown at the BOTTOM-LEFT corner.\\
- The pyramid defines the TARGET orientation:\\
  - The pyramid’s tip indicates the FORWARD direction.\\
  - The pyramid implicitly defines Yaw, Pitch, and Roll.\\

\#\# IMAGE 2 (Generated Image)\\
- The model-generated image after orientation editing.\\

Your task is to evaluate whether the FINAL orientation in IMAGE 2 matches the TARGET orientation defined by the pyramid in IMAGE 1.\\

------------------------------------------------------------\\
\#\# STEP 1: Identify target object and orientation anchor\\

Identify the main target object whose orientation is edited.\\

Then determine its Orientation Anchor — the semantic part that defines the object’s forward-facing direction.\\

Use these rules:\\
- Shoes → toe\\
- Chairs → backrest\\
- Cars → front / headlights\\
- Cameras → lens\\
- Humans → face\\
- Animals → head\\
- Tools / devices → functional front end\\

State explicitly:\\
- Target object\\
- Orientation anchor\\

------------------------------------------------------------\\
\#\# STEP 2: Describe TARGET orientation (IMAGE 1 ONLY)\\

Look ONLY at IMAGE 1.\\

Describe the pyramid’s orientation using three independent axes:\\

- Yaw: horizontal facing direction of the pyramid’s tip\\
- Pitch: upward or downward tilt of the pyramid’s tip\\
- Roll: leftward or rightward tilt of the pyramid body\\

Rules:\\
- Do NOT reference the object.\\
- Do NOT compare with IMAGE 2.\\

\end{tcolorbox}
\label{tab:app_prompt_reo_oa1}
\end{table*}

\begin{table*}[]
\centering
\caption{Evaluation Prompt for Orientation Alignment 2/3}
\begin{tcolorbox}[title = {Orientation Alignment 2/3}]

- Do NOT use vague terms.\\
- Use explicit spatial descriptions.\\

------------------------------------------------------------\\
\#\# STEP 3: Describe FINAL object orientation (IMAGE 2 ONLY)\\

Look ONLY at IMAGE 2.\\

Using the orientation anchor, describe the FINAL orientation of the object in IMAGE 2:\\

- Yaw\\
- Pitch\\
- Roll\\

Rules:\\
- Describe the actual observed orientation, not intent.\\
- Do NOT reference IMAGE 1 or the pyramid.\\
- Do NOT infer past states.\\
- Do NOT merge axes.\\
- If an axis cannot be determined, mark it as N/A.\\

------------------------------------------------------------\\
\#\# STEP 4: Axis-wise alignment (comparison)\\

Now compare the descriptions from STEP 2 and STEP 3.\\

For EACH axis:\\

- Score = 1\\
  if the FINAL object orientation matches the TARGET pyramid orientation.\\

- Score = 0\\
  otherwise.\\

IMPORTANT:\\
- Scoring is based ONLY on final alignment.\\
- Do NOT use any information about whether modification was needed.\\

------------------------------------------------------------\\
\#\# STEP 5: Needs\_Modification analysis (IMAGE 1 ONLY — report only)\\

Look ONLY at IMAGE 1 again.\\

For each axis, decide whether the object’s ORIGINAL orientation matched the pyramid’s TARGET orientation:\\

- Needs\_Modification = YES\\
- Needs\_Modification = NO\\

Rules:\\
- This step is for reporting only.\\
- It MUST NOT affect the score.\\

\end{tcolorbox}
\label{tab:app_prompt_reo_oa2}
\end{table*}

\begin{table*}[]
\centering
\caption{Evaluation Prompt for Orientation Alignment 2/3}
\begin{tcolorbox}[title = {Orientation Alignment 2/3}]

- Do NOT reference IMAGE 2.\\
- Do NOT infer from the final result.\\

------------------------------------------------------------\\
\#\# OUTPUT FORMAT\\

First provide reasoning under the following sections:\\

- \#\# Object \& Anchor\\
- \#\# Target Orientation (IMAGE 1)\\
- \#\# Final Orientation (IMAGE 2)\\
- \#\# Axis-wise Alignment\\
- \#\# Needs Modification (Report Only)\\

Then output the final result as JSON (LAST PART ONLY):\\

\{\\
\ \ ``Yaw'': \{\\
\ \ \ \ ``needs\_modification'': true,\\
\ \ \ \ ``reason'': ``string'',\\
\ \ \ \ ``score'': 0\\
\ \ \},\\
\ \ ``Pitch'': \{\\
\ \ \ \ ``needs\_modification'': false,\\
\ \ \ \ ``reason'': ``string'',\\
\ \ \ \ ``score'': 0\\
\ \ \},\\
\ \ ``Roll'': \{\\
\ \ \ \ ``needs\_modification'': true,\\
\ \ \ \ ``reason'': ``string'',\\
\ \ \ \ ``score'': 0\\
\ \ \}\\
\}\\

\end{tcolorbox}
\label{tab:app_prompt_reo_oa3}
\end{table*}

\begin{table*}[]
\centering
\caption{Evaluation Prompt for Reorientation Contextual Preservation 1/2}
\begin{tcolorbox}[title = { Reorientation Contextual Preservation 1/2}]
You are given TWO images.\\

\#\# IMAGE 1 (Input Image)\\
- The original image before editing.\\

\#\# IMAGE 2 (Generated Image)\\
- The model-generated image after orientation editing.\\

Your task is to evaluate TWO independent metrics:\\

1. Identity Consistency (IC)\\
2. Visual Integrity (VI)\\

These two metrics MUST be evaluated separately and independently.\\

------------------------------------------------------------\\
\#\# PART 1: Identity Consistency (IC)\\

Identity Consistency evaluates whether the edited subject in IMAGE 2 remains the same semantic object/entity as in IMAGE 1, ignoring changes that are directly caused by orientation editing.\\

\#\#\# STEP 1: Identify the edited subject\\

- Look at IMAGE 1 and identify the main subject intended to be edited.\\
- Look at IMAGE 2 and identify the corresponding subject.\\

\#\#\# STEP 2: Identity consistency check\\

Decide whether the subject in IMAGE 2 is the same identity as in IMAGE 1.\\

You MUST ignore:\\
- Orientation or facing-direction changes.\\
- Pose changes caused by reorientation.\\
- Perspective changes due to rotation.\\
- Lighting, shading, or color changes.\\
- Minor texture or detail variations.\\

You MUST penalize:\\
- Object replacement (e.g., shoe → different shoe type or different object).\\
- Object removal or duplication.\\
- Structural corruption (missing parts, extra parts, broken anatomy).\\
- Conversion into a fundamentally different semantic entity.\\

\#\#\# IC Scoring (binary):\\
- Score = 1\\
  if the subject identity and overall structure are clearly preserved.\\

- Score = 0\\
  if the subject is replaced, missing, duplicated, or structurally corrupted.\\

\end{tcolorbox}
\label{tab:app_prompt_reo_rcp1}
\end{table*}

\begin{table*}[]
\centering
\caption{Evaluation Prompt for Reorientation Contextual Preservation 1/2}
\begin{tcolorbox}[title = { Reorientation Contextual Preservation 1/2}]

If unsure, score 0.\\

------------------------------------------------------------\\
\#\# PART 2: Visual Integrity (VI)\\

Visual Integrity evaluates whether the editing process introduced severe visual artifacts or layout corruption that significantly degrade image quality.\\

\#\#\# STEP 3: Visual corruption inspection\\

Compare IMAGE 2 against IMAGE 1 and check for severe issues such as:\\

- Prominent UI or indicator pollution (e.g., orientation pyramid or markers appearing in the image center).\\
- Large unnatural visual artifacts or rendering glitches.\\
- Structural image collapse (e.g., extreme distortion, tearing, duplicated geometry).\\
- Occlusions or obstructions that severely harm image readability.\\

You MUST ignore:\\
- Orientation changes of the edited subject.\\
- Normal perspective changes caused by rotation.\\
- Lighting or color changes.\\
- Minor artifacts that do not significantly affect readability.\\

\#\#\# VI Scoring (binary):\\
- Score = 1\\
  if IMAGE 2 remains visually coherent and readable, with no severe artifacts or layout corruption.\\

- Score = 0\\
  if IMAGE 2 exhibits clear visual breakdown or UI pollution.\\

If unsure, score 0.\\

------------------------------------------------------------\\
\#\# OUTPUT FORMAT\\

First provide your reasoning under:\\
- \#\# Identity Consistency Analysis\\
- \#\# Visual Integrity Analysis\\

Then output the final result as JSON (LAST PART ONLY):\\

\{
\ \ ``Identity\_Consistency'': \{\\
\ \ \ \ ``reason'': ``string (one-sentence summary)'',\\
\ \ \ \ ``score'': 0\\
\ \ \},\\
\ \ ``Visual\_Integrity'': \{\\
\ \ \ \ ``reason'': ``string (one-sentence summary)'',\\
\ \ \ \ ``score'': 0\\
\ \ \}
\}\\

\end{tcolorbox}
\label{tab:app_prompt_reo_rcp}
\end{table*}


\end{document}